\documentclass[letterpaper]{article}
\def\year{2018}\relax

\usepackage{aaai18}  %Required
\usepackage{times}  %Required
\usepackage{helvet}  %Required
\usepackage{courier}  %Required
\usepackage{url}  %Required
\usepackage{graphicx}  %Required
\PassOptionsToPackage{usenames,dvipsnames}{xcolor}
\usepackage{wrapfig}

\usepackage{framed}
\usepackage{amssymb}
\usepackage{amsfonts}
\usepackage{amsmath} 

\usepackage[colorlinks=false]{hyperref}
\usepackage{paralist}

\usepackage{array}
\usepackage{booktabs}
\usepackage{todonotes}
\usepackage[caption=false]{subfig}
\usepackage{wrapfig}
\usepackage[sort&compress]{natbib}
\bibpunct{[}{]}{,}{n}{}{,}   % numbered
\usepackage{xcolor}
\usepackage{xspace,hyperref}
\usepackage{multirow}

\newcommand{\toolname}{{\sc DPDebugger}\xspace}
\newcommand{\kernelname}{{\sc Alignment}\xspace}

\usepackage{listings}

\usepackage{color}

\definecolor{dkgreen}{rgb}{0,0.6,0}
\definecolor{gray}{rgb}{0.5,0.5,0.5}
\definecolor{mauve}{rgb}{0.58,0,0.82}

\makeatletter
\newif\if@restonecol
\makeatother

\usepackage[ruled,vlined,linesnumbered,lined]{algorithm2e}
\usepackage{algpseudocode}

\newcommand\vx{\mathbf{x}}
\newcommand\vy{\mathbf{y}}
\newcommand\vz{\mathbf{z}}

\newcommand\vX{\mathbf{X}}

\newcommand\vZ{\mathbf{Z}}

\newcommand\va{\mathbf{A}}

\newcommand\vc{\mathbf{c}}

\usepackage{fancyvrb}

\newcommand{\Ee}{{\cal E}}
\newcommand{\Ff}{{\cal F}}
\newcommand{\Tt}{{\cal T}}

\newcommand\tupleof[1]{\left\langle #1 \right\rangle}

\newcommand{\set}[1]{\left\{ #1 \right\}}

\newcommand{\seq}[1]{\langle #1 \rangle}
\newcommand{\Nat}{\mathbb N}

\newcommand{\R}{\mathbb R}

\newcommand{\Real}{\R}

\usepackage{adjustbox}

\def\rmdef{\stackrel{\mbox{\rm {\tiny def}}}{=}} % equals definition

\definecolor{gold}{rgb}{0.99,0.78,0.07}
\usetikzlibrary{arrows,shapes,snakes,automata,backgrounds,positioning,decorations.pathmorphing,
	decorations.markings,calc}

\tikzstyle{dtreenode}=[draw=blue!10!gray,rounded rectangle, minimum size=5mm,fill=blue!10!white]
\tikzstyle{dtreeleaf}=[draw=black!60,minimum width=1cm,minimum height=0.4cm,rectangle,fill=blue!50!white]
\tikzset{every loop/.style={looseness=7}}
\tikzset{
	gluon/.style={decorate,draw=black,
		decoration={coil,amplitude=1pt, segment length=5pt}} 
}
\tikzset{
	gluon1/.style={decorate,draw=black,
		decoration={coil,amplitude=3pt, segment length=3pt}} 
}
\tikzset{
	gluonew/.style={decorate,draw=black,
		decoration={coil,amplitude=1pt, segment length=2pt}} 
}

\tikzset{bicolor/.style args={#1 and #2 and #3}{
		path picture={
			\tikzset{rounded corners=0}
			\fill [#1] (path picture bounding box.south west)
			rectangle
			($(path picture  bounding box.north west)!#3!(path picture bounding
			box.north east)$);
			\fill [#2]
			($(path picture bounding box.south west)!#3!(path picture bounding
			box.south east)$)
			rectangle (path picture bounding box.north east);
}}}

\tikzset{tricolor/.style args={#1 and #2 and #3 and #4 and #5}{
		path picture={
			\tikzset{rounded corners=0}
			\fill [#1] (path picture bounding box.south west)
			rectangle
			($(path picture  bounding box.north west)!#4!(path picture bounding
			box.north east)$);
			\fill [#2]
			($(path picture bounding box.south west)!#4!(path picture bounding
			box.south east)$)
			rectangle
			($(path picture  bounding box.north west)!#5!(path picture bounding
			box.north east)$);
			\fill [#3]
			($(path picture bounding box.south west)!#5!(path picture bounding
			box.south east)$)
			rectangle (path picture bounding box.north east);
}}}

\definecolor{cadmiumgreen}{rgb}{0.0, 0.42, 0.24}

\definecolor{verde}{rgb}{0.25,0.5,0.35}
\definecolor{jpurple}{rgb}{0.5,0,0.35}
\definecolor{darkgreen}{rgb}{0.0, 0.2, 0.13}

\newcommand{\estiloJava}{
	\lstset{
		language=Java,
		basicstyle=\ttfamily\small,
		keywordstyle=\color{jpurple}\bfseries,
		stringstyle=\color{red},
		commentstyle=\color{verde},
		morecomment=[s][\color{blue}]{/**}{*/},
		extendedchars=true,
		showspaces=false,
		showstringspaces=false,
		numbers=left,
		numberstyle=\tiny,
		breaklines=true,
		backgroundcolor=\color{cyan!10},
		breakautoindent=true,
		captionpos=b,
		xleftmargin=0pt,
		tabsize=2
}}

\usepackage[inline]{enumitem}

\usepackage{filecontents,pgfplots}
\usepackage{wrapfig}

\begin{document}
	
	\title{Differential Performance Debugging with Discriminant Regression Trees
		\thanks{This research was supported in part by DARPA under agreement FA8750-15-2-0096.}}
	\author{
		Saeid Tizpaz-Niari,
		Pavol {\v C}ern\'y,
		Bor-Yuh Evan Chang,
		Ashutosh Trivedi \\
		University of Colorado Boulder  
	}
	\maketitle
	\begin{abstract}
		Differential performance debugging is a technique to
		find performance problems.
		It applies in situations where the performance of a
		program is (unexpectedly) different for
		varying classes of inputs.
		The task is to explain the differences in asymptotic
		performance among various input classes
		in terms of program internals.
		We propose a data-driven technique based on
		{\em discriminant regression tree} (DRT) learning problem where
		the goal is to discriminate among different classes
		of inputs.
		We propose a new algorithm for DRT learning that first clusters the data into functional clusters, capturing different asymptotic performance classes, and then invokes off-the-shelf decision tree learning algorithms to explain these clusters.
		We focus on linear functional clusters and adapt classical clustering
		algorithms ($K$-means and spectral) to produce them.
		For the $K$-means algorithm, we generalize the notion of the cluster centroid from a point to a linear function. We
		adapt spectral clustering by defining a novel kernel function to
		capture the notion of ``linear'' similarity between two data points.
		We evaluate our approach on benchmarks consisting of Java programs
		where we are interested in debugging performance.
		We show that our algorithm outperforms other well-known
		regression tree learning algorithms in terms of running
		time and accuracy of classification.
	\end{abstract}
	
	\section{Introduction}
	\label{sec:introduction}
	Developers often face the problem of finding and fixing performance
	problem in their programs.
	Performance bugs manifest themselves only on certain pathological inputs.
	For instance, there can be two inputs of the
	same size on which the performance is
	unexpectedly different in an otherwise functionally correct
	program.
	
	We study the {\em differential performance
		problem}, where the goal
	is to explain the
	difference in performance between two classes of inputs in terms of
	program internals, such as which functions were called and how many
	times were they called. This information is useful, as it allows a
	programmer or an analyst to better assess whether the performance
	difference is inherent to the problem, or is a result of a coding
	inadequacy. The problem is hard for both
	traditional static as well as dynamic analysis
	techniques. Static analysis commonly target logical correctness
	properties (and not the performance), and are not as scalable as
	techniques based on machine learning.
	On the other hand, dynamic analysis techniques
	such as profiling, focus on individual traces, whereas for
	the differential performance problem, we need to compare the performance
	on different traces.
	
	We propose a technique called {\em differential performance
		debugging}, based on inference of {\em discriminant regression
		trees} (DRTs). DRTs are regression trees where the goal is to classify
	input data. In contrast, the objective of standard regression tree
	learning is to predict the output for a previously unseen input.
	The input to the differential performance
	problem is a set of program traces. Each trace is represented as
	follows. We have input variables (such as the size of the
	user input), auxiliary variables (such as the functions called), and the
	output variable (such as the running time).
	The output to the differential performance
	problem is the DRT. The internal nodes of the tree has
	predicates on auxiliary variables. The leaf nodes
	model the output variable as a function of input
	variables. The leaf nodes
	represent the performance for different classes of inputs capturing asymptotically
	different performance behaviors.
	
	In accordance with Occam's razor, we are interested in finding a DRT with a small number of
	clusters, while minimizing the modeling error. Furthermore, the DRT
	should be a human readable explanation, which also suggests that
	smaller number of clusters is preferable.
	There are two major steps in our algorithm.
	First, we project the data into the input and output variables
	and cluster the data in this domain. Second, we consider
	the auxiliary variables only and identify what
	separates the clusters in terms of these variables. We use an
	off-the-shelf decision tree learning algorithm for
	the second step. The first, clustering, step thus reduces the regression tree
	inference problem to the decision tree inference problem.
	
	For our approach, we need a clustering algorithm that produces {\em
		functional} clusters, that is, clusters that represent
	functions from input variables to the output variable. We adapt two classical clustering
	algorithms. First, we extend the $K$-means algorithm to produce linear
	functional clusters. This is done by generalizing the notion
	of the cluster centroid from a point to a linear function. Second, we adapt the spectral clustering
	algorithm by defining a new notion of similarity between two data
	points that we dub {\em alignment} kernel. Here, two data points are more similar when the line defined by them
	captures more data points.

	The key contributions of this paper are:
	\begin{compactitem}
		\item We propose {\em discriminant regression trees} which are
		regression trees where the goal is to classify input data into a
		small number of clusters.
		\item We give a new algorithm for learning discriminant regression
		trees. It finds (functional) clusters first, which enables learning
		the tree using an efficient algorithm for learning decision trees.
		\item We present extensions to two classical clustering algorithms:
		$k$-means and spectral clustering. These
		extensions allow us to obtain functional clusters.
		\item We implement our approach in the tool \toolname and evaluate it
		on benchmarks consisting of a suite of Java programs.
		Our experiments that the approach is scalable and is able to explain the
		differences in performance between different classes of inputs.
	\end{compactitem}

	\section{Overview}
	\label{sec:overview}
	We show how our prototype tool
	\toolname can be used for diagnosing performance problems on a
	real-life example. We also use the example to explain how the tool
	works and compare it to existing approaches.

	\noindent\textbf{Performance problem with Apache FOP.}
	Apache FOP (Formatting Objects Processor) is
	a Java application that reads a formatting object such as an XML file
	and renders the resulting pages
	to a specified output format such as PDF and PS.
	The formatting document can specify that
	an external image in, for example, a PNG or JPEG format should be
	included.
	A user had a suspicion that there is a performance bug in handling PNG
	images. They reported in a forum
	post in $2011$
	that they have two PNG
	images, which have the same size, but one of them takes seven times as much to
	render as the other one\footnote{https://bz.apache.org/bugzilla/show\_bug.cgi?id=51465}.

	\noindent\textbf{Performance debugging with \toolname.}
	Our tool, \toolname, can be used exactly in this situation, to
	help an analyst to explain the differences in performance. The analyst
	can then decide whether the differences are inherent to the problem or
	they are a manifestation of a coding error.
	Song and Lu \cite{song2014statistical} reported that in 60\% of bugs analyzed by
	them, users notice huge performance differences among similar inputs.
	
	The analyst has to collect a number of inputs which in this case are
	PNG and JPEG images of various sizes. We remark
	that in many cases, the inputs can be collected from log files of a system or generated by
	existing software fuzzers~\cite{cadar2008klee}.
	
\begin{figure}[!t]
		\caption{\footnotesize Performance clusters (left) in the FOP data set and a
			DRT (right) explaining the the clusters in terms of function calls.
		}
		\label{fig:FOP}
		\scalebox{0.45}{
			\begin{tikzpicture}
			\begin{axis}[xlabel=Image size (in bytes),ylabel= Time (s)]
			\addplot[
			visualization depends on={\thisrow{id}\as\myvalue},
			scatter/classes={
				0={mark=+,blue},
				1={mark=+,red}
			},
			scatter, only marks,
			scatter src=explicit symbolic]
			table[x=size,y=mean,meta=label]
			{data/fop.dat};
			\end{axis}
			\end{tikzpicture}
		}
		\hfill
		\scalebox{0.45}{
			\begin{tikzpicture}[align=center,node distance=0.8cm,->,thick,
			draw = black!60, fill=black!60]
			\centering
			\pgfsetarrowsend{latex}
			\pgfsetlinewidth{0.3ex}
			\pgfpathmoveto{\pgfpointorigin}
			\pgfusepath{stroke}
			
			\node[dtreenode,initial above,initial text={}] at (0,0) (l0)  {
				ps.ImageEncodingHelper.\\encodeRenderImageWith-\\DirectColorModelAsRGB};
			\node[dtreenode,below right=of l0] (l2)
			{profile.\\ColorProfileUtil.\\
				getICC\_Profile};
			\node[dtreenode,below left=of l2] (l3) {inline.Line\\LayoutManager\\.handleOverflow};
			
			\node[dtreeleaf,bicolor={red and red and 0.99},below left=of l0] (l1) {};
			\node[dtreeleaf,bicolor={blue and red and 0.99},below right=of l2]  (l4) {};
			\node[dtreeleaf,bicolor={red and blue and 0.85},below left=of l3] (l5) {};
			\node[dtreeleaf,bicolor={blue and red and 0.9},below right=of l3]  (l6) {};
			
			\path[->]  (l0) edge  node [left,pos=0.4] {$< 1 ~~$} (l1);
			\path  (l0) edge  node [right, pos=0.4] {$~~ \geq 1$} (l2);
			\path  (l2) edge  node [left] {$\leq 1 ~~$} (l3);
			\path  (l2) edge  node [right] {$~~ > 1$} (l4);
			\path  (l3) edge  node [left] {$=0 ~~$} (l5);
			\path  (l3) edge  node [right] {$~~ =1$} (l6);
			\end{tikzpicture}
		}
	\end{figure}
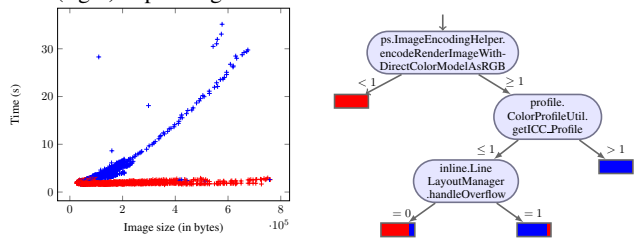

	Given the collection of inputs, the tool produces the two diagrams in
	Figure~\ref{fig:FOP}. The analyst can diagnose the performance problem
	using these two figures. The left diagram is a plot of the image size (input variable) and the running time (output variable). From the plot,
	the analyst can see that there are two performance clusters.
	However, the analyst does not know what separates these two clusters. It
	is instructive to emphasize that looking at the two groups of inputs
	does not explain the difference. All the JPEG images are in the lower
	(red) cluster, but there are PNG images of similar size in both clusters.
	
	We thus turn to the DRT on the right side of Figure~\ref{fig:FOP}
	for the explanation. It says that if for an input,
	the function {\tt encodeRender...RGB} is not called, then the input will be in the red cluster. The user can
	analyze the reverse call graph to see how the function is called, and realize that it is
	called for PNG files, but not for JPEG files. Further, the node to the
	right of the root has the function {\tt getICCprofile}. This function is
	what distinguishes the red (fast) cluster from the blue (slow) one among PNG
	files. It is called once for every PNG file, but it is called more
	than once only for PNG files with a color scheme that needs to be deflated. After code
	analysis, we see that one source
	of the performance problem is that some PNG files have a compressed
	color scheme that needs to be deflated. Another way how a PNG image can be in the blue cluster
	is that the dimension of the input image overflows the allowed size
	(see the lowest internal node).
	
	The user thus learns from the discriminant regression tree (DRT) that what
	separates the two clusters is the fact that the images in the blue
	clusters either need to be deflated first, or overflow. So in this case, the diagnosis is
	that the difference in performance is not a coding mistake but it is inherent
	to the problem. The user can conclude this without needing to sift through
	almost 40,000 methods that Apache FOP has.
	
	\noindent\textbf{Inside the tool \toolname.}
	We now describe how \toolname obtains the diagrams in
	Figure~\ref{fig:FOP}. The diagram on the left side is obtained as
	follows.
	The program is run on all the inputs, and the graph of
	input size versus running time is plotted. Then we need to cluster the
	data. As a preliminary step, we need to
	get the values of the auxiliary variables. In this example, they
	indicate how many times a function was called. We have a variable for
	each (non-standard library) function.
	Then, we find functional clusters in the data from the left diagram. That is we
	consider input and output variables only (not auxiliary variables) and
	identify a given number $K$ of clusters. These clusters are intended
	to capture the classes of inputs with different (asymptotic)
	performance. For our example, spectral clustering identifies the
	clusters marked red and blue in Figure~\ref{fig:FOP}.
	
	To obtain the right diagram, the next step is decision tree learning. We want to learn what
	separates the clusters in terms of predicates on the auxiliary
	variables. Note that here each data point is labeled with one of the
	$K$ labels (indicating to which cluster the data point belongs),
	therefore efficient decision tree learning (such as CART) can now be used to
	construct the regression tree. The final discriminant regression tree is on the right
	part of Figure~\ref{fig:FOP}.

	\noindent\textbf{Comparison with existing regression-tree algorithms.}
	We applied the state-of-the-art algorithms for learning regression trees
	(M5Prime~\cite{witten2016data} and GUIDE~\cite{loh2002regression}) to our problem.
	Our
	goal is different from the goal of these algorithms: we aim to classify data, whereas both Guide and
	M5Prime aim to predict the output for previously unseen input. We
	believe that this accounts for the following differences.
	
	M5Prime finds a linear regression tree with 23 different linear models
	in leafs. Guide finds 4 linear models -- two of them same as our
	algorithm, but two of the in-between, perhaps to account for
	noise. Note that \toolname identified 2 clusters.
	Furthermore, the running time of the algorithms are as follows: M5Prime 97 seconds,
	Guide 1233.6 seconds, and our \toolname 14.4 seconds.

	\section{Discriminant Regression Tree Learning}
	\label{sec:problem}
	Let $\vX = \set{\vx_1, \vx_2, \ldots, \vx_n} \in \Real^n$ be the set of input
	variables, $\vZ = \set{\vz_1, \ldots, \vz_m} \in \Real^m$ be the
	set of auxiliary variables,  and $\vy \in \Real$ be the
	performance (observable output) variable of our target  program. 
	To keep the presentation simple, we assume that there is a single performance
	variable $\vy$, although techniques presented in the paper can easily be
	extended to include a set of performance variables representing time-series data
	on various performance measures such as time and memory.
	
	An {\it execution trace} $T$ of the program is a tuple $\tupleof{X, Z, y}$ wherein
	$X = \seq{x_1, x_2, \ldots, x_n}$, $Z = \set{z_1, \ldots, z_m}$, and $y$
	represent the valuations to the input, auxiliary, and output variables,
	respectively. 
	We further assume that the valuations of the auxiliary variables deterministically
	depend only on valuations of the input variables.
	However, we allow the process of measuring performance to be noisy.
	Due to this we can potentially have multiple traces of the program with the  
	same values for the input and the auxiliary variables but different values for the
	performance variable.
	
	A \emph{trace discriminant} is defined as a disjoint hyper-rectangular
	partitioning of the space of auxiliary variables along with 
	an affine function for each partition modeling the performance variable as a
	function only of input 
	variables.
	Formally, a trace discriminant $\Psi = (\Ff, P)$ is a set of affine
	functions $\Ff = \seq{f_1, f_2, \ldots, f_K}$---where each $f_j
	: \Real^n \to \Real$ models the performance variable $y$ as a
	function of the input variables---and a hyper-rectangular partition
	$P = \seq{(\phi_1, d_1), (\phi_2, d_2), \ldots, (\phi_l, d_l)}$ where each
	$\phi_i\::\: \Real^m \to \set{\texttt{true}, \texttt{false}}$  is a
	hyper-rectangular predicate over auxiliary variables $\vZ$, and each 
	$d_i: \Ff \to [0, 1]$ is discrete probability distribution over $\Ff$ giving a
	probabilistically weighted modeling of the affine functions from $\Ff$.
	The size $\texttt{size}(\Psi)$ of a discriminant $\Psi$ is defined as total
	number of affine functions (i.e., $\texttt{size}(\Psi) = |\Ff|$).
	Given a trace $T = \seq{X, Z, y}$ and a discriminant $\Psi = (\Ff, P)$, we
	define the prediction error  as
	%\[
	$\epsilon(T, \Psi)  \rmdef   \left(y - \sum_{j=1}^{K} d_i(f_j) \cdot f_j(x_1,
	x_2, \ldots, x_n)\right)^2$,          
	%\]
	where $1 {\leq} i {\leq} l$ is the index of the unique partition in $P$ such that
	$Z \models \phi_i$ (i.e., the predicate $\phi_i$ evaluates to true for the valuation $Z$).
	Given a set of traces $\Tt = \set{T_1, T_2, \ldots, T_N}$, and a discriminant
	$\Psi$, we define the fitness of the discriminant as {\emph
		mean-squared-error}
	%\[
	$\mu(\Tt, \Psi) = \frac{1}{N} \sum_{i=1}^{N} \epsilon(T_i, \Psi)$.
	%\]
	\begin{algorithm}[bp!]\normalsize
		{\scriptsize 
			\DontPrintSemicolon
			\KwIn{A set of traces $\Tt = \set{T_1, T_2, \ldots, T_N}$, an upper bound
				$B$ on discriminant size, and a bound on mean-squared error $B_\varepsilon$.}
			\KwOut{ Return a discriminant regression tree of size $B$ and error bound
				$B_\varepsilon$, if possible. 
				Otherwise  return $\textsc{Null}$.} 
			
			Extract points $\Ee = \set{\seq{X_1, y_1}, \seq{X_2, y_2}, \ldots, \seq{X_N,
					y_N}}$ from the trace set $\Tt = \set{T_1, \ldots, T_N}$ where
			$T_i = \seq{X_i, Z_i, y_i} \in \Tt$.
			
			Using linear clustering algorithms presented in the next section, find the
			smallest number $B'$ of linear clusters $\Ff = \set{f_1, \ldots, f_{|B'|}}$
			that can fit the data with mean-squared error smaller than $B_\varepsilon$.\;
			
			\lIf {$B' > B$}{  \Return \textsc{Null}}
			\Else
			{
				Extract points $\Ee' = \set{\seq{Z_1, \ell_1}, \ldots, \seq{Z_N, \ell_N}}$
				where $\ell_i \in \Ff$ is the label assigned by the clustering algorithm
				to the dataset $\seq{X_i, y_i}$.
				
				Use a standard decision tree algorithm to learn a decision tree (along
				with its accuracy based on $k$-fold cross-validation) from $\Ee'$.
				
				Return the discriminant regression tree (along with its accuracy) by
				replacing labels at the leaves with corresponding linear functions. 
			}
			
			\caption{\scriptsize \textsc{LearnDiscriminantRegressionTree}$(\Tt, B, B_\varepsilon)$}
			\label{Alg:RegTreeLearn}
		}
	\end{algorithm}
	
	Given a set of traces $\Tt$, a bound on the size of the discriminant
	$B \in \Nat$, and a bound on the error $B_\varepsilon \in \Real$, the
	\emph{discriminant learning problem} is to find a discriminant $\Psi$ with
	$\texttt{size}(\Psi) \leq B$ and $\mu(\Tt, \Psi) \leq B_\varepsilon$.
	It follows from Theorem 1 in~\cite{AS14} that the discriminant learning problem
	is \textsc{NP-hard}.
	For this reason, we use heuristics to construct discriminant using 
	classification and regression trees.
	
	A {\it discriminant regression tree} is a trace discriminant represented as a
	binary tree structure whose nodes contain predicates over auxiliary variables and leaves
	contain a discrete probability distribution over affine functions in input variables.
	An example of a distribution regression tree is shown in
	Figure~\ref{fig:FOP} where each leaf represents a partition, and the probability
	distribution over functions is pictorially depicted using relative sizes of
	different colors.  
	
	Classical regression tree algorithms can be used to learn the discriminant
	regression trees.
	The most straightforward way to generalize the decision tree algorithm to learn
	regression trees is computationally expensive~\cite{loh2011} as it 
	requires solving two linear regression problems for each split candidate.
	Popular regression tree algorithms algorithms
	CART~\cite{breiman1984classification}, M5Prime~\cite{witten2016data},
	GUIDE~\cite{loh2002regression} propose various ways to avoid this problem. 
	CART is a piecewise constant regression tree model that
	uses the standard regression-tree algorithm (with piecewise constant clusters)
	and then applies cross-validation to prune the tree. 
	M5Prime~\cite{witten2016data} algorithm first constructs a piecewise constant
	model, and then fits linear regression models to leaves during pruning
	step.
	GUIDE regression tree algorithm~\cite{loh2002regression}, at each node, fits the
	best regression model that predicts the response variable and computes the residual.
	Then, it adds different class labels for traces with negative and positive
	residuals and solves classification problem to find the
	auxiliary variable to split over.
	
	In our setting (where the goal is classification and a tight upper bound on the number of linear
	clusters is known), we propose a simple but rather effective method to overcome
	the complexity of repeatedly fitting the piecewise linear model.
	Our approach is summarized as Algorithm~\ref{Alg:RegTreeLearn}.
	Our approach is to first cluster traces along the lines based
	only on input and output variables, and then assign different labels to various
	traces based on the linear clusters into which they fall.
	The next step is to learn a classification decision tree in auxiliary variables
	with the leaves as clusters labels (classes) learned in the first step.
	Using a set of microbenchmarks related to performance debugging, in
	Section~\ref{sec:experiment} we show that
	our algorithm performs better than other regression tree algorithms specially
	when the relationship between auxiliary variables and linear clusters is complex.
	
	\section{Linear Clustering Algorithms}
	\label{sec:clustering}
	In this section, we study $K$-linear clustering problem required at the
	clustering step of Algorithm~\ref{Alg:RegTreeLearn}.
	Consider the set $\Ee = \set{\seq{X_1, y_1}, \ldots, \seq{X_N, y_N}}$ of
	data points where $X_i \in \Real^n$ is an $n$-dimensional vector of valuations
	to the input variables and $y_i \in \Real$ is the value of the output (response)
	variable. 
	Given the number of desired clusters $K$, the {\it $K$-linear clustering problem}
	asks to compute a partition of set $\Ee$ into $K$ clusters
	$S=\seq{S_1, S_2, \ldots, S_K}$  minimizing the  residual-sum-of-squares (RSS) 
	defined as 
	%\[
	%RSS(S) :=
	$\sum_{i=1}^{K} \sum_{\seq{X, y} \in S_i} \min_{f \in \Ff}
	\left(y - f(X)\right)^2$,
	%\]
	where $f$ is a linear function $f : \Real^n \to \Real$ over
	input variables in the form of $f(\vx) {=} \va \vx {+} \vc$ with
	$\vx$ as an $n$-dim. vector and $\vc$ as a scalar.
	
	Regarding the computational complexity of the {\it $K$-linear clustering problem},
	observe that for a given cluster $S$, the RSS can be computed using (least
	squares) linear regression in polynomial time (linear in the number of points
	and quadratic in input dimension). Since, there are only finitely many ($K^N$)
	distinct clusters possible, the $K$-linear clustering problem is decidable.  
	The NP-hardness of $K$-linear clustering problem follows from NP-hardness of
	$K$-means clustering problem which is known to be NP-hard both for general
	dimensions and $2$ clusters~\cite{DDHP09}, as well as $2$ dimension and $K$
	clusters~\cite{MPV09}.
	For this reason, we present two heuristics to solve $K$-linear clustering
	problem. 
	The first algorithm, which we call ``K-linear'' clustering, extends $K$-means
	algorithm by using line centroids instead of point centroids,
	while the second algorithm is
	based on spectral clustering with a new notion of measuring similarity between
	points in order to detect functional relationships.  
	
	\subsection{K-Linear Clustering}
	We propose a modification of the standard $K$-means clustering
	algorithm to give a heuristic to solve $K$-linear clustering problem
	as shown in Algorithm~\ref{Klinear}.
	The termination of our algorithm is guaranteed as the number of distinct
	clusters possible are finite, and in each step we get a strict improvement
	in residual-sum-of-squares due to the restriction~(\ref{eqq1}) in
	Algorithm~\ref{Klinear} on changing the set only in the case of
	a strict improvement.
	However, similar to the $K$-means algorithm, there is
	no guaranteed convergence to a global optimum.
	The choice of initial partition to fit linear ``centroids'' is crucial in
	converging towards the global optimal solution.
	One way to choose a good partition is to pick lines defined by pairs of points
	such that $\epsilon$-size tubes around the lines pass through a large number of 
	points. 
	Another possible heuristic to achieve better partition is similar to that often
	seen with $K$-means algorithm---we execute the $K$-linear algorithm a couple of
	times with randomly selected initial partitions, and then we choose the result that gives minimum RSS. 
	
	\begin{algorithm}[tbp!]\normalsize
		{\scriptsize 
			\DontPrintSemicolon
			\KwIn{Data $\Ee = \set{\seq{X_1, y_1}, \ldots, \seq{X_N, y_N}}$ and
				number of clusters $K$.}
			\KwOut{A partition of the set of traces $\Ee$ in $K$ sets
				$\seq{S_1, S_2, \ldots, S_K}$.} 
			Let $S^{(1)} = \seq{S_1^{(1)}, S_2^{(1)}, \ldots, S_K^{(1)}}$ be an arbitrary
			partition of the points $\Ee$.
			
			Set $i$ to $0$\;
			\Repeat{$S^{(i)} = S^{(i+1)}$}
			{
				Set $i$ to $i+1$\;
				For each set $S_j^{(i)} = \set{\seq{X_P, y_P}}$ where $\seq{X_P, y_P}$ is
				a set of points assigned to partition $S_j$ at $i$-th iteration,
				learn a linear function $f_j = \va_j \vx + c_j$ minimizing
				$\sum_{p \in P} (y_p - \va_j X_p - c_j)^2$.
				
				Compute  $S^{(i+1)} = \seq{S_1^{(i+1)}, S_2^{(i+1)}, \ldots, S_K^{(i+1)}}$
				such that for each $1 \leq j \leq k$  we have
				$\seq{X_P, y_P} \in S_j^{(i+1)}$ if 
				$(y_p {-} \va_j X_p {-} c_j)^2 = \min_{1 \leq h \leq k} (y_p {-} \va_h X_p {-} c_h )^2$,
				with condition that $\seq{X_p, y_p} {\in} S_j^{(i)}$ and
				$\seq{X_p, y_p} {\in} S_h^{(i+1)}$ for $j{\not =} h$ implies
				\begin{equation}
					(y_p - \va_j X_p - c_j)^2 > (y_p - \va_h X_p - c_h)^2\label{eqq1}
				\end{equation}
			}
			\Return clusters $S^{(i)}$ and linear ``centroids'' $\seq{f_1, f_2,
				\ldots, f_K}$.
			\caption{\scriptsize \textsc{$K$-LinearClusteringAlgorithm}}
			\label{Klinear}
		}
	\end{algorithm}

	\subsection{Spectral Clustering with \kernelname Kernel}
	Spectral clustering is a popular clustering algorithm that views the
	clustering data as a weighted graph of points and the clustering problem as
	a graph partitioning problem.
	Spectral clustering algorithms are parameterized by the notion of adjacency between
	two data points defined using {\it kernel functions}.
	Spectral clustering is useful in clustering problems where the
	measurement of the center and the spread of cluster are not a suitable description of
	clusters~\cite{von2007tutorial}. 
	
	In order to define the notion of adjacency in terms of being close to a given
	linear cluster, we characterize a novel kernel function---called {\it alignment
		kernel}---that puts two points closer to each other if the line passing through
	those points have multiple other points in the line's neighborhood.
	\begin{wrapfigure}{l}{0.5\columnwidth}
		\resizebox{0.5\columnwidth}{!}{
			\begin{tikzpicture}
			\draw[style=help lines,step=0.5cm] (0,0) grid (6.2,6.2);
			\draw[->,thick] (-0.1,0) -- (6.5,0) node[anchor=west]{x};
			\draw[->,thick] (0,-0.1) -- (0,6.5) node[anchor=south]{y};
			
			\foreach \x in {0,1,...,6} \draw [thick](\x cm,-2pt) -- (\x cm,2pt);
			\foreach \y in {0,1,...,6} \draw [thick](-2pt,\y) -- (2pt,\y);
			\foreach \x in {0,1,...,6} \draw (\x cm, 0 cm) node[anchor=north]{\x};
			\foreach \y in {0,1,...,6}  \draw (0 cm, \y cm) node[anchor=east]{\y};
			
			\begin{scope}[color=black]
			\filldraw (1,1) circle (0.08cm) node (A) {}
			node[anchor=north,fill=white,yshift=-0.1cm] {$A$};
			\filldraw (6,6) circle (0.08cm) node (B) {}
			node[anchor=west,fill=white,xshift=5pt] {$B$};
			\filldraw (3, 1) circle (0.08cm) node (C) {}
			node[below,anchor=south,fill=white,yshift=0.1cm] {$C$};
			\end{scope}
			
			\draw [ultra thick, draw=yellow!10, fill=yellow, opacity=0.2]
			(0.75,1.25) -- (5.75,6.25) -- (6.25,5.75) -- (1.25, 0.75) -- cycle;
			
			\draw [ultra thick, draw=pink!10, fill=pink, opacity=0.2]
			(1,1.35) -- (5.5,1.35) -- (5.5,0.65) -- (1, 0.65) -- cycle;
			
			\begin{scope}[color=black!30]
			\filldraw (1.5,1.8) circle (0.08cm) node (ab1) {};
			\filldraw (2.5,2.3) circle (0.08cm) node (ab2) {};
			\filldraw (3,3) circle (0.08cm) node (ab3) {};
			\filldraw (4, 4.2) circle (0.08cm) node (ab4) {};
			\filldraw (4, 3.8) circle (0.08cm) node (ab5) {};
			\filldraw (5,5.3) circle (0.08cm) node (ab6) {};
			\filldraw (6,5.8) circle (0.08cm) node (ab7) {};
			\filldraw (5.5, 3.5) circle (0.08cm) node (ab8) {};
			\filldraw (5, 3) circle (0.08cm) node (ab9) {};
			\filldraw (2.5,2.7) circle (0.08cm) node (ab10) {};
			\filldraw (5, 5.3) circle (0.08cm) node (ab11) {};
			\filldraw (4, 2) circle (0.08cm) node (ab12) {};
			\filldraw (4, 0.8) circle (0.08cm) node (ab13) {};
			\end{scope}
			
			\draw[very thick] (A) -- (B) node[pos=0.5, xshift=1cm, yshift=1cm, sloped,
			below] {$\bf \alpha(A, B) = 1/2^8$};
			\draw[very thick] (A) -- (5.5,1) node[pos=0.8, yshift=-0.2cm, below] {$\bf \alpha(A, C) = 1/2$};
			\end{tikzpicture}
		}
	\end{wrapfigure}
	The concept of alignment kernel is shown in the figure on the left 
	where points $A$ and $B$ are closer to each-other in linear sense than points
	$A$ and $C$, although the latter points are closer than former points in terms of
	Euclidean distance.
	Given a data set $\Ee$, we define an \emph{alignment kernel function}
	$\alpha^\Delta_\Ee: (\Real^n\times\Real) \times (\Real^n \times\Real) \to \Real$
	to be a real-valued symmetric and non-negative function defined as the following
	for every pair of neighboring points $\seq{X_i,y_i}$ and $\seq{X_j,y_j}$:
	\begin{eqnarray*}
		\alpha^\Delta_\Ee(\seq{X_i,y_i},\seq{X_j,y_j}) \rmdef \begin{cases}
			0 & \text{ if $i = j$}\\
			2^{-|R_{ij}|} & \text{if $|R_{ij}| >= 1$}\\ 
			\infty & \text{otherwise,}
		\end{cases}
	\end{eqnarray*}
	where $R_{ij}$ is the set of points $\seq{X_r, y_r}$ in $\Ee$ such that $r
	{\neq} i {\neq} j$, and it has $\Delta$ distance from the line passing through 
	$\seq{X_i,y_i}$ and $\seq{X_j,y_j}$.
	Finally, we construct similarity matrix by calculating
	$\mathrm{e}^{-\alpha^\Delta_\Ee(A, B)}$ for every pair of points $A$ and $B$.
	Observe that the exact computation of the alignment matrix is cubic in number of
	data points. However, we have implemented a quadratic procedure (see
	supplemental material in \cite{1711.04076}) computing an approximation of the the alignment kernel.  
	In our experiments, we have found that for linear clusters the quality of the
	alignment kernel is better than the RBF and the nearest-neighborhood kernels.
	In comparison with the $K$-linear clustering, spectral clustering with alignment
	kernel can often detect non-linear clusters.

	\section{Microbenchmark Results}
	\label{sec:experiment}
	\begin{table*}[tbp!]
		\caption{Micro-benchmark results for comparison different affine cost model learning algorithms.
			Legend: \textbf{\#M}: number of functions, \textbf{\#N:} number of traces,
			\textbf{T}:\ computation time in seconds, \textbf{$R^2$}: coefficient of determination
			\textbf{H}: decision-tree height,
			\textbf{L}: Number of detected models,
			\textbf{A}: accuracy of classification model, \textbf{$\epsilon < 0.1$ sec.}
		}
		\label{tab:reg-vs-class}
		\resizebox{\textwidth}{!}{
			\begin{tabular}{ || l | r | r || r | r | r | r || r| r | r | r ||r| r | r |
					r || r| r | r | r | r ||r| r | r | r | r ||}
				\hline
				&       &    & \multicolumn{4}{c||}{CART} & \multicolumn{4}{c||}{M5prime}& \multicolumn{4}{c||}{GUIDE} & \multicolumn{5}{c||}{\toolname ($K$-linear)}& \multicolumn{5}{c||}{\toolname (spectral)} \\
				\cline{4-25}
				Bench& \textbf{\#M} & \textbf{\#N}
				& \textbf{T} & \textbf{$R^2$} &  \textbf{H} & \textbf{L}
				& \textbf{T} & \textbf{$R^2$} &  \textbf{H} & \textbf{L}
				& \textbf{T} & \textbf{$R^2$} &  \textbf{H} & \textbf{L}
				& \textbf{T} & \textbf{A} & \textbf{$R^2$} &  \textbf{H} & \textbf{L}
				& \textbf{T} & \textbf{A} & \textbf{$R^2$} &  \textbf{H} & \textbf{L}\\ \hline
				R\_2 & 2 & 400 & $\epsilon$  & 0.99 & 14 & 237 & 3.5 & 0.99 & 6 & 17 &
				$\epsilon$ & 0.99 & 2 & 4
				& 0.7 & 99\% & 0.99 & 2 & 3 & 0.2 & 96\% & 0.98 & 2 & 3 \\ \hline
				R\_3\#1 & 3 & 800 & 0.15 & 0.77 & 14 & 486 & 4.5 & 0.7 & 1 & 1 &
				$\epsilon$ & 0.99 & 3 & 8
				& 1.3 & 100\% & 0.99 & 3 & 2 & 0.7 & 100\% & 0.99 & 3 & 2 \\ \hline
				R\_3\#2 & 3 & 800 & 0.14 & 0.99 & 15 & 470 & 6.9 & 0.9 & 8 & 41 &
				$\epsilon$ & 0.86 & 3 & 6
				& 1.7 & 100\% & 0.99 & 3 & 3 & 0.8 & 99\% & 0.99 & 3 & 3 \\ \hline
				R\_4\#2 & 4 & 1200 & 0.2 & 0.99 & 14 & 652 & 8.8 & 0.99 & 7 & 23 &
				0.2 & 0.99 & 3 & 5
				& 2.9 & 100\% & 0.99 & 4 & 4 & 1.6 & 98\% & 0.99 & 4 & 4 \\ \hline
				R\_4\#1 & 4 & 1600 & 0.28 & 0.99 & 20 & 893 & 9.7 & 0.99 & 7 & 25 &
				0.2 & 0.99 & 4 & 7
				& 3.5 & 99\% & 0.99 & 4 & 3 & 3.0 & 98\% & 0.99 & 4 & 3 \\ \hline
				R\_4\#3 & 4 & 1600 & 0.27 & 0.97 & 16 & 955 & 9.9 & 0.9 & 8 & 87 &
				0.2 & 0.93 & 4 & 11
				& 3.5 & 99\% & 0.99 & 4 & 3 & 2.8 &  99\% & 0.99 & 4 & 3 \\ \hline
				R\_5 & 5 & 3200 & 0.54 & 0.94 & 16 & 1810 & 17.6 & 0.71 & 11 & 147 & 0.4
				& 0.73 & 5 & 15
				& 6.7 & 99\% & 0.99 & 5 & 3 & 11.3 & 99\% & 0.99 & 5 & 3 \\ \hline
				R\_6 & 6 & 6400 & 1.1 & 0.99 & 22 & 3695 & 24.1 & 0.98 & 12 & 173 & 1.1
				& 0.6
				& 4 & 11
				& 16.3 & 99\% & 0.99 & 6 & 4 & 42.5 & 98\% & 0.99 & 6 & 4 \\ \hline
				R\_7 & 7 & 12800 & 2.4 & 0.99 & 32 & 5126 & 49.6 & 0.99 & 12 & 142 & 1.4 & 0.63 & 4 & 14 & 31.8 & 97.9\% & 0.98 & 7 & 4 & 210.1 & 95.5\% & 0.97 & 7 & 4 \\ \hline
			\end{tabular}
		}
		\vspace{-2em}
	\end{table*}
	\begin{table}[!t]
		\begin{center}
			\caption{Micro-benchmark results comparing GUIDE and \toolname with dummy function calls.
			}
			\label{tab:reg-vs-class-2}
			\resizebox{\columnwidth}{!}{
				\begin{tabular}{ || l | r | r || r| r | r | r ||r| r | r | r | r ||}
					\hline
					&       &    & \multicolumn{4}{c||}{GUIDE}& \multicolumn{5}{c||}{\toolname ($K$-linear)} \\
					\cline{4-12}
					Benchmark& \# \textbf{M} & \#\textbf{N}
					& \textbf{T} & \textbf{$R^2$} &  \textbf{H} & \textbf{L}
					& \textbf{T} & \textbf{A} & \textbf{$R^2$} & \textbf{H} & \textbf{L}\\ \hline
					R\_200 & 200 & 400 & 3.0 & 0.99 & 2 & 4 & 0.7 & $99.6\%$ & 0.99 & 2 & 3 \\ \hline
					R\_400\#1 & 400 & 800 & 12.0 & 0.99 & 4 & 12 & 1.3 & $95.0\%$ & 0.99 & 3 & 2 \\ \hline
					R\_400\#2 & 400 & 800 & 12.2 & 0.99 & 3 & 8 & 1.8 & $100\%$ &0.99 & 3 & 3 \\ \hline
					R\_600 & 600& 1200 & 38.2 & 0.99 & 3 & 5 & 3.4 & $100\%$  & 0.99 & 4 & 4 \\ \hline
					R\_800\#1 & 800 & 1600 & 85.0 & 0.99 & 4 & 7 &  3.9 & $98.9\%$ & 0.99 & 4 & 3 \\ \hline
					R\_800\#2 & 800 & 1600 & 84.6 & 0.96 & 4 & 15 & 4.0 & $98.3\%$ & 0.99 & 4 & 3 \\ \hline
					R\_1600 & 1600 & 3200 & 624.9 & 0.73 & 4 & 15 & 9.1 & $98.3\%$ & 0.99 & 5 & 3 \\ \hline
					R\_3200 & 3200 & 6400 & 4706.2 & 0.59 & 4 & 12 & 59.5 & $99.1\%$ & 0.99 & 6 & 4 \\ \hline
					R\_6400 & 6400 & 12800 & 34897.0 & 0.63 & 4 & 14 & 261.2 & $98.0\%$ & 0.99 & 7 & 4 \\ \hline
				\end{tabular}
			}
		\end{center}
		\vspace{-2em}
	\end{table}
	
	\noindent\textbf{Empirical evaluation questions.}
	We compare our approach to existing algorithms for learning regression
	trees and focus on the following questions.
	\begin{enumerate*}
		\item {\it How deep are the regression trees and how many leaves do they have?} As
		we expect our regression trees to serve as human readable
		explanations, smaller height and number of leaves are better.
		\item
		{\it How
			scalable is our approach compared to the existing approaches?}
		\item
		{\it What is the prediction ability (as measured by coefficient of
			determination) compared to state-of-the-art approaches?} The metric we
		are interested in is accuracy of classification based on $10$-fold
		cross-validation. However, the standard approaches are built
		with the goal of prediction measured by coefficient of determination
		$R^2$. We therefore compare performances with $R^2$.
	\end{enumerate*}
	
	\noindent\textbf{Synthetic benchmarks.} We compared the performance of
	regression tree learning approaches on a set of microbenchmarks.   The benchmarks were constructed
	in such a way that the clusters have increasingly complex characterizations. 
	We consider micro-benchmarks named R\_n\#v where n is the  number of functions
	in the benchmark, and v is the version number. For
	R\_2 to R\_7, in each case there is a cluster which consists of inputs where all
	the functions were called. R\_200 to R\_6400 are versions of the same benchmarks
	with many other functions that do not influence performance. Each function call
	executes a for loop statement where the number of iterations depends on the
	inputs. Each trace of a benchmark invokes a set of functions that lead to the
	different time of execution. We want to detect clusters and explain them based
	on function calls.
	
	\noindent\textbf{Results.}
	Table~\ref{tab:reg-vs-class} summarizes the results of applying
	CART, M5Prime, and GUIDE (the standard algorithms) as well as our algorithms
	with $K$-linear and spectral clustering.
	
	Our first question is about the simplicity of the explanation produced by our
	tool.
	Table~\ref{tab:reg-vs-class} shows that CART and M5Prime produce
	significantly deeper trees with more models than GUIDE and
	\toolname. For instance, for benchmark R\_7, CART produces a tree with
	more than 5000 nodes, M5Prime with more than one node, whereas the
	benchmark has only 4 clusters.
	
	Our second question is about scalability. As the first question
	established that CART and M5Prime are unsuitable for our purpose, we
	compare scalability only with the GUIDE algorithm.
	Table~\ref{tab:reg-vs-class-2} shows the performance of
	GUIDE regression tree and \toolname with $K$-linear clustering
	when there are many features (function calls). We see that \toolname
	is more scalable on this set of benchmarks. For instance, for R\_6400,
	GUIDE takes more than 9 hours, whereas \toolname takes less than 5
	minutes.
	
	Our third question asks to compare the coefficient of
	determination $R^2$. From Table~\ref{tab:reg-vs-class}, we have the
	following: CART and M5Prime generally perform well (but there are some
	outliers where the coefficient of determination drops). The main
	problem with these algorithms for our purpose is the large size of the
	regression trees. For GUIDE, the coefficient of determination is
	lower for more complex examples. \toolname performs uniformly well in
	this metric.
	
	Finally, we compare the two versions of our algorithm: one with
	$K$-linear and one with spectral
	clustering. Table~\ref{tab:reg-vs-class} shows that for these
	benchmarks with linear clusters, they are similar in all metrics except
	running time, where the $K$-linear clustering is slightly better.

	\section{Case Study}
	\label{sec:case-study}
	\begin{table}[tbp]
		\begin{center}
			\caption{Java applications studied using \toolname{}.}
			\label{tab:apps-case-study}
			\resizebox{\columnwidth}{!}
			{        
				\begin{tabular}{ c|  c | c }
					\hline
					{\LARGE Name} & {\LARGE Cluster} & {\LARGE Discriminant Regression Tree}\\ \hline
					{\rotatebox[origin=c]{90}{\LARGE\textsc     Charts4j~~}} &
					\adjustbox{valign=t}{
						\begin{tikzpicture}
						\begin{axis}[xlabel=Number of data points,ylabel=Time (s)]
						\addplot[
						visualization depends on={\thisrow{id}\as\myvalue},
						scatter/classes={
							0={mark=+,blue},
							1={mark=+,red},
							2={mark=+,green!40!black}
						},
						scatter, only marks,
						scatter src=explicit symbolic]
						table[x=size,y=mean,meta=label]
						{data/charts4j.dat};
						\end{axis}
						\end{tikzpicture}
					}
					&
					\adjustbox{valign=t}{
						\begin{tikzpicture}[align=center,node distance=2.2cm,->,thick, draw = black!60, fill=black!60]
						\pgfsetarrowsend{latex}
						\pgfsetlinewidth{0.3ex}
						\pgfpathmoveto{\pgfpointorigin} 
						\pgfusepath{stroke}
						
						\node[dtreenode,initial above,initial text={}] at (0,0) (l0)  {charts4j.\\DataUtil.\\Scale};
						\node[dtreenode,below left=of l0] (l1)  {charts4j.\\PlotImpl.\\$<$init$>$};
						\node[dtreeleaf,bicolor={cadmiumgreen and blue and .9},
						below right=of l0] (l2) {};
						\node[dtreeleaf,color={blue},below left=of l1]  (l3) {};
						\node[dtreeleaf,tricolor={blue and red and cadmiumgreen and .3 and .9},
						below right=of l1]  (l4) {};
						
						\path  (l0) edge  node [left] {$= 0~~~$} (l1);
						\path  (l0) edge  node [right] {$~~~= 1$} (l2);
						\path  (l1) edge  node [left] {$= 0~~~$} (l3);
						\path  (l1) edge  node [right] {$~~~= 1$} (l4);
						\end{tikzpicture}
					}
					\\ \hline
					{\rotatebox[origin=c]{90}{\LARGE\textsc     Snapbuddy~~}} &
					\adjustbox{valign=t}{
						\begin{tikzpicture}
						\begin{axis}[xlabel=Image size (in bytes),ylabel= Time (s)]
						\addplot[
						visualization depends on={\thisrow{id}\as\myvalue},
						scatter/classes={
							0={mark=+,cyan},
							1={mark=+,black},
							2={mark=+,red},
							3={mark=+,blue},
							4={mark=+,green!40!black}
						},
						scatter, only marks,
						scatter src=explicit symbolic]
						table[x=size,y=mean,meta=label]
						{data/SB.dat};
						\end{axis}
						\end{tikzpicture}
					}
					&
					\adjustbox{valign=t}{
						\begin{tikzpicture}[align=center,node distance=1cm,->,thick,
						draw = black!60, fill=black!60]
						\pgfsetarrowsend{latex}
						\pgfsetlinewidth{0.3ex}
						\pgfpathmoveto{\pgfpointorigin} 
						\pgfusepath{stroke}
						
						\node[dtreenode,initial above,initial text={}] at (0,0) (l0)  {Image.\\~~~KaleidoscopeFilter.filter~~~};
						\node[dtreenode,below left=of l0] (l1) {model.\\filter.filter};
						\node[dtreenode,below right=of l0] (l2) {Image.\\~~~TritoneFilter.filter~~~};
						\node[dtreeleaf,bicolor={red and cadmiumgreen and 0.9},below left=of l1] (l3) {};
						\node[dtreenode,below=of l1] (l4)
						{image.\\EmbossFilter};
						\node[dtreenode,below left=of l2] (l5)
						{image.\\EmbossFilter};
						\node[dtreenode,below=3.1cm of l2] (l6)
						{image.\\EmbossFilter};
						\node[dtreeleaf,bicolor={cadmiumgreen and cyan and 0.9},below left=of l4] (l7) {};
						\node[dtreenode,below=of l4] (l8)
						{image.\\TritoneFilter.\\filter};
						\node[dtreeleaf,bicolor={black and blue and 0.8},below=of l5] (l9) {};
						\node[dtreeleaf,bicolor={cyan and blue and 0.8},below right=of l5] (l10) {};
						\node[dtreeleaf,bicolor={cyan and blue and 0.9},below left=of l6] (l11) {};
						\node[dtreeleaf,bicolor={blue and cyan and 0.95},below =of l6] (l12) {};
						\node[dtreeleaf,bicolor={cadmiumgreen and cyan and 0.9},below left=of l8] (l13) {};
						\node[dtreeleaf,bicolor={black and blue and 0.95},below right=of l8] (l14) {};	
						
						\path  (l0) edge  node [left] {$False ~~$} (l1);
						\path  (l0) edge  node [right] {$~~ True$} (l2);
						\path  (l1) edge  node [left] {$False ~~$} (l3);
						\path  (l1) edge  node [right] {$~~ True$} (l4);
						\path  (l2) edge  node [left] {$False ~~$} (l5);
						\path  (l2) edge  node [right] {$~~ True$} (l6);
						\path  (l4) edge  node [left] {$False ~~$} (l7);
						\path  (l4) edge  node [right] {$~~ True$} (l8);
						\path  (l5) edge  node [left] {$False ~~$} (l9);
						\path  (l5) edge  node [right] {$~~ True$} (l10);
						\path  (l6) edge  node [left] {$False ~~$} (l11);
						\path  (l6) edge  node [right] {$~~ True$} (l12);
						\path  (l8) edge  node [left] {$False ~~$} (l13);
						\path  (l8) edge  node [right] {$~~ True$} (l14);
						\end{tikzpicture}
					}
					\\ \hline
					{\rotatebox[origin=c]{90}{\LARGE\textsc     JFreeChart~~}} &    
					\adjustbox{valign=t}{
						\begin{tikzpicture}
						\begin{axis}[xlabel=Number of data points,ylabel=
						Time (s)]
						\addplot[
						visualization depends on={\thisrow{id}\as\myvalue},
						scatter/classes={
							0={mark=+,blue},
							1={mark=+,red}
						},
						scatter, only marks,
						scatter src=explicit symbolic]
						table[x=size,y=mean,meta=label]
						{data/jfreechart.dat};
						\end{axis}
						\end{tikzpicture}
					}
					&
					\adjustbox{valign=t}{
						\begin{tikzpicture}[align=center,node distance=3.2cm,->,thick, draw = black!60, fill=black!60]
						\pgfsetarrowsend{latex}
						\pgfsetlinewidth{0.3ex}
						\pgfpathmoveto{\pgfpointorigin} 
						\pgfusepath{stroke}
						
						\node[dtreenode,initial above,initial text={}] at (0,0) (l0)  {xy.Candlestick\\Renderer.drawItem};
						\node[dtreeleaf,bicolor={blue and red and .97},
						below left=of l0] (l1) {};
						\node[dtreeleaf,bicolor={red and blue and .95},
						below right=of l0] (l2) {};
						
						\path  (l0) edge  node [left] {$= 0 ~~$} (l1);
						\path  (l0) edge  node [right] {$~~> 0$} (l2);
						\end{tikzpicture}
					}
					\\ \hline
					{\rotatebox[origin=c]{90}{\LARGE\textsc     Apache POI~~}} &    
					\adjustbox{valign=t}{
						\begin{tikzpicture}
						\begin{axis}[xlabel=Size of slides in Byte,ylabel= Time (s)]
						\addplot[
						visualization depends on={\thisrow{id}\as\myvalue},
						scatter/classes={
							0={mark=+,blue},
							1={mark=+,red},
							2={mark=+,green!40!black}
						},
						scatter, only marks,
						scatter src=explicit symbolic]
						table[x=size,y=mean,meta=label]
						{data/POI.dat};
						\end{axis}
						\end{tikzpicture}
					}
					&
					\adjustbox{valign=t}{
						\begin{tikzpicture}[align=center,node distance=2.2cm,->,thick, draw = black!60, fill=black!60]
						\pgfsetarrowsend{latex}
						\pgfsetlinewidth{0.3ex}
						\pgfpathmoveto{\pgfpointorigin} 
						\pgfusepath{stroke}
						
						\node[dtreenode,initial above,initial text={}] at (0,0) (l0)  {hslf.blip.\\PNG.getType};
						\node[dtreeleaf,color={blue}, below left=of l0] (l1) {};
						\node[dtreenode,below right=of l0] (l2) {ddf.EscherSimple\\Property.getPropertyValue};
						\node[dtreeleaf,bicolor={green!40!black and blue and .7},below left=of l2] (l3) {};
						\node[dtreeleaf,color={red},below right=of l2] (l4) {};
						
						\path  (l0) edge  node [left] {$= 0 ~~$} (l1);
						\path  (l0) edge  node [right] {$~~> 0$} (l2);
						\path  (l2) edge  node [left] {$<= 20 ~~$} (l3);
						\path  (l2) edge  node [right] {$~~> 20$} (l4);
						\end{tikzpicture}
					}
					\\\hline
					{\rotatebox[origin=c]{90}{\LARGE\textsc     Ode4j~~}} &
					\adjustbox{valign=t}{
						\begin{tikzpicture}
						\begin{axis}[xlabel=Number of geom objects,ylabel=
						Time (s)]
						\addplot[
						visualization depends on={\thisrow{id}\as\myvalue},
						scatter/classes={
							0={mark=+,blue},
							1={mark=+,red}
						},
						scatter, only marks,
						scatter src=explicit symbolic]
						table[x=size,y=mean,meta=label]
						{data/ode4j.dat};
						\end{axis}
						\end{tikzpicture}
					}
					&
					\adjustbox{valign=t}{
						\begin{tikzpicture}[align=center,node distance=3.2cm,->,thick, draw = black!60, fill=black!60]
						\pgfsetarrowsend{latex}
						\pgfsetlinewidth{0.3ex}
						\pgfpathmoveto{\pgfpointorigin} 
						\pgfusepath{stroke}
						
						\node[dtreenode,initial above,initial text={}] at (0,0) (l0)  {internal.DxSAP.\\Space.collide};
						\node[dtreeleaf,bicolor={blue and red and 0.97}, below left=of l0] (l1) {};
						\node[dtreeleaf,bicolor={red and blue and 0.7},below right=of l0] (l2) {};
						
						\path  (l0) edge  node [left] {$= 0 ~~$} (l1);
						\path  (l0) edge  node [right] {$~~> 0$} (l2);
						\end{tikzpicture}
					}
					\\\hline
					{\rotatebox[origin=c]{90}{\LARGE\textsc     Collab~~}} &
					\adjustbox{valign=t}{
						\begin{tikzpicture}
						\begin{axis}[xlabel=Number of ADD operations,ylabel=
						Time (s)]
						\addplot[
						visualization depends on={\thisrow{id}\as\myvalue},
						scatter/classes={
							0={mark=+,blue},
							1={mark=+,red}
						},
						scatter, only marks,
						scatter src=explicit symbolic]
						table[x=size,y=mean,meta=label]
						{data/collab.dat};
						\end{axis}
						\end{tikzpicture}
					}
					&
					\adjustbox{valign=t}{
						\begin{tikzpicture}[align=center,node distance=3.2cm,->,thick, draw = black!60, fill=black!60]
						\pgfsetarrowsend{latex}
						\pgfsetlinewidth{0.3ex}
						\pgfpathmoveto{\pgfpointorigin} 
						\pgfusepath{stroke}
						
						\node[dtreenode,initial above,initial text={}] at (0,0) (l0)  {collab.Scheduling\\Sandbox.split};
						\node[dtreeleaf,bicolor={blue and red and 0.97}, below left=of l0] (l1) {};
						\node[dtreeleaf,bicolor={red and blue and 0.98}, below right=of l0] (l2) {};
						\path  (l0) edge  node [left] {$<= 2805~~$} (l1);
						\path  (l0) edge  node [right] {$~~> 2805$} (l2);
						\end{tikzpicture}
					}
					\\\hline
					{\rotatebox[origin=c]{90}{\LARGE\textsc     Tweeter~~}} &
					\adjustbox{valign=t}{
						\begin{tikzpicture}
						\begin{axis}[xlabel=Number of words in a tweet,ylabel=
						Time (s)]
						\addplot[
						visualization depends on={\thisrow{id}\as\myvalue},
						scatter/classes={
							0={mark=+,blue},
							1={mark=+,red}
						},
						scatter, only marks,
						scatter src=explicit symbolic]
						table[x=size,y=mean,meta=label]
						{data/tweeter.dat};
						\end{axis}
						\end{tikzpicture}
					}
					&
					\adjustbox{valign=t}{
						\begin{tikzpicture}[align=center,node distance=3.2cm,->,thick, draw = black!60, fill=black!60]
						\pgfsetarrowsend{latex}
						\pgfsetlinewidth{0.3ex}
						\pgfpathmoveto{\pgfpointorigin} 
						\pgfusepath{stroke}
						
						\node[dtreenode,initial above,initial text={}] at (0,0) (l0)  {hibernate.Alternative\\Suggestions.\\getAlternatives};
						\node[dtreeleaf,color={blue}, below left=of l0] (l1) {};
						\node[dtreeleaf,color={red},below right=of l0] (l2) {};
						
						\path  (l0) edge  node [left] {$= 0 ~~$} (l1);
						\path  (l0) edge  node [right] {$~~> 0$} (l2);
						\end{tikzpicture}
					}
					\\\hline
				\end{tabular}
			}
		\end{center}
	\end{table}
	
	Table~\ref{tab:case-study} summarizes the results over eight case studies
	involving large Java applications, while
	Table~\ref{tab:apps-case-study} shows clustering and classification
	steps in \toolname for seven of them (the eighth is in Figure~\ref{fig:FOP}).
	We next explain how \toolname explained differences in performance for different classes of inputs for these seven applications 
	(Apache FOP has been treated already in Section~\ref{sec:overview})~\footnote{see
		supplemental material for detailed information in \cite{1711.04076}}.
	\begin{enumerate}
		\item
		{\bf Charts4j}.
		Charts4j is a Java chart library that enables developers
		to generate different charts available in the Google Chart API.
		The input data set for our experiments consists of asking for
		different plots of the same data.
		Our tool applied spectral clustering with the number of clusters set to $3$.
		It finds three functional clusters between execution time
		and the number of data items.
		The DRT for this experiment
		(shown in Table~\ref{tab:apps-case-study}) gives two key insights:
		First, we notice that green cluster consists of plots that call {\it scale}
		function.
		Upon further investigation, the scale function is indeed expensive since it
		computes minimum and maximum for the input array and normalized
		values for each element of the input array.
		Next, the class initialization for \texttt{PlotImpl} distinguishes blue and red
		clusters. This happens for plots with one dimension
		where it needs to generate another dimension.
		It also needs to convert the new data set to a double array.
		This part is a performance bug because it can build a
		double array structure for the new dimension
		instead of double list.
		
		\item {\bf Snapbuddy}.
		SnapBuddy is a mock social network application where users can make their public
		profiles~\cite{borges2017model,tizpaz2017discriminating}.
		As inputs for our experiments, we passively monitored  users' interaction with public
		profile pages.
		We applied $K$-linear clustering algorithm where we set the number of clusters to
		$5$.
		As a result,  our tool finds five linear relationships between the size of public
		profile image and download time.
		The DRT learned by our tool is shown in Table~\ref{tab:apps-case-study}.
		It reports that filter combinations applied in profile pictures are the key discriminants.
		
		\item {\bf JFreeChart}.
		JFreeChart is a free Java chart library that
		helps developers to plot different charts in their applications.
		The input data set is the set of open-high-low-close (OHLC) items.
		We use this library to plot OHLC items with different renders.
		Our tool applied $K$-linear clustering algorithm with $K=2$.
		The clustering step finds two linear relationships between
		time of execution and the number of OHLC items.
		The DRT produced by our algorithm is shown in Table~\ref{tab:apps-case-study}.
		We observe that whether the plot applies
		\texttt{CandlestickRenderer} and calls \texttt{drawItem} distinguishes
		between red and blue clusters.
		In particular, the number of
		calls to \texttt{drawItem} is equal to the number of data points in OHLC dataset, and each call executes
		the loop statement when a candlestick renderer is applied. This performance bug
		was also reported in~\cite{olivo2015static} using static analysis.
		
		\item {\bf Apache POI}.
		The Apache POI Project's mission is to read and write MS Excel, MS Word, and
		PowerPoint files using Java.
		As inputs to our experiments, we used different slides that
		randomly include text, table, images, shapes, and smartArts,
		We applied the spectral clustering algorithm with number of clusters set to
		3.
		The DRT produced by our algorithm is shown in Table~\ref{tab:apps-case-study}.
		The DRT shows that all program traces labeled with the blue cluster do not have
		any image data source (like PNG, smartArts, and so on). Green and red clusters
		are distinguished only based on their item sizes.
		
		\item {\bf Ode4j}.
		Ode4j is an open source, high performance library for simulating rigid body
		dynamics.
		The data set includes the different number of geom objects that interact with
		different APIs.
		We applied the spectral clustering algorithm with the number of clusters set to $2$.
		The clustering finds two functional relationships between the time of execution
		and number of geom objects.
		The DRT (Table~\ref{tab:apps-case-study}) reports \texttt{collide} function in
		\texttt{DxSAPSpace} class  to distinguish between blue and red clusters.
		Looking into the source code, we can see the quadratic behavior in
		\texttt{collide} function. As there is a linear cost function named
		\texttt{collide2} in \texttt{DxSAPSpace} class, we suspect that quadratic
		behavior can be mitigated.
		
		\item {\bf Collab}.
		Collab is a scheduling application that allows
		users to create new events and modify existing
		events. The data set consists of different operations
		to add and modify events using \texttt{ADD}, \texttt{UNDO},
		and \texttt{DONE} commands.
		We applied  $K$-linear clustering algorithm with $K$ equals to $2$.
		It finds two linear relationships between the time of execution and the number of
		add operations.
		The DRT produced by our algorithm is shown in Table~\ref{tab:apps-case-study}.
		The discriminant model shows that the traces in red cluster
		call \texttt{split} function more times than traces in
		blue cluster. The average number of calls to
		\texttt{split} function for blue cluster is $1,678$ and for
		red cluster is $88,507$.
		The temporary store data structure
		of Collab is a B-tree. The number of times that the B-tree needs
		to split a parent node is the main explanation to distinguish
		red and blue clusters.
		
		\item {\bf Tweeter}.
		Tweeter application is a mock of Twitter application.
		Users can post tweets and see tweets posted by other users.
		The data set consists of a maximum $20$ words for a tweet.
		We applied $K$-linear clustering algorithm where we set $K$ to $2$.
		The clustering step finds two relationships
		between time of execution and number of words in a tweet.
		The DRT produced by our algorithm is shown in Table~\ref{tab:apps-case-study}.
		The discriminant model produces \texttt{getAlternatives}
		function as a feature to distinguish blue and red clusters.
		This function is called more times when the input tweet
		includes mistakes, and it will never be called if the tweet
		is correct.
	\end{enumerate}
	
	\begin{table}[!t]
		\begin{center}
			\caption{Java applications studied using \toolname.
				Legend: \textbf{\#$M_T$}: total number of functions in application,
				\textbf{\#$M_R$}: total number of observed functions,
				\textbf{\#N:} number of collected traces,
				\textbf{$C_T$:} Clustering algorithm ($K$ for $K$-linear and $S$ for spectral),
				\textbf{T}:\ computation time of \toolname in seconds,
				\textbf{A}: accuracy of classification model,
				\textbf{H}: decision-tree height,
				\textbf{\#C}: Number of clusters,
				\textbf{$\epsilon < 0.1$ sec.}
			}
			\label{tab:case-study}
			\resizebox{\columnwidth}{!}{
				\begin{tabular}{ || l | r | r | r | r | r | r | r | r ||}
					\hline
					Application & \# \textbf{$M_T$} & \#\textbf{$M_R$} & \#\textbf{N}
					& \textbf{$C_T$} & \textbf{T} & \textbf{A}
					& \textbf{H} & \# \textbf{C} \\ \hline
					Apache FOP & 39,694 & 2,765 & 1,988  & $S$ & 14.4 & 96.4\%
					& 3 & 2 \\ \hline
					Charts4j & 715 & 71 & 2,000 & $S$ &11.7 & 87.1\%
					& 2 & 3 \\ \hline
					SnapBuddy & 3,071 & 150 & 616  & $K$ & 1.6 & 88.6\%
					& 5 & 4 \\ \hline
					JFreeChart & 9,162 & 527 & 1,000 & $K$ &1.6 & 99.4\%
					& 1 & 2 \\ \hline
					Apache POI & 10,396  & 199 & 661 & $S$ & 0.44 & 86.4\%
					& 2 & 3 \\ \hline
					Ode4j & 4,564 & 114 & 577 & $S$ & 7.8 & 85.0\%
					& 1 & 2 \\ \hline
					Collab & 185 & 53 & 530 & $K$ & 0.7 & 96.6\% & 1 & 2 \\ \hline
					Tweeter & 947 & 31 & 320 & $K$ & 0.6 &  100\% & 1 & 2 \\ \hline
				\end{tabular}
			}
		\end{center}
		\vspace{-2em}
	\end{table}

	\section{Related Works}
	\label{sec:related}
	\paragraph{Performance debugging}
	The work by Tizpaz-Niari et al.~\cite{tizpaz2017discriminating}
	is the closest to ours. They use decision tree learning for finding
	security vulnerabilities, whereas we focus on performance bugs.
	They heavily rely on the assumption that it is enough to consider
	constant clusters, whereas we consider a much more realistic and
	the general setting of linear functional clusters. On the algorithmic
	side, for Tizpaz-Niari et al. it is sufficient to use the standard
	$K$-means algorithm (since they have only constant clusters), whereas
	we needed to adapt the $K$-means and clustering algorithms for the
	linear functional case.
	
	Spectrum-based fault localization is often used for explaining
	bugs~\cite{wong2016survey,song2014statistical, jin2010instrumentation,
		jones2002visualization,liblit2003bug}. In particular,
	Song and Lu~\cite{song2014statistical} use a statistical based technique
	for performance debugging problem.
	They indicate that the problems are manifested by the
	performance not being uniform, but rather there being
	two set of inputs: bad and good inputs.
	They refer to bug report databases to
	generate the bad inputs. They use statistical models to find
	predicates that distinguish good and bad
	inputs.
	They assume that two sets of bad and good
	inputs are given. In contrast, we are not given any
	labeled data set. We obtain different classes by clustering techniques. In addition, we
	are not limited to two sets of inputs.
	Finally, Song and Lu find the ranking of
	predicates and choose the top one that is responsible for
	bad runs. On the other hand, we use decision tree
	learning that can produce conjunctions of
	predicates.
	This can be seen in the SnapBuddy example.
	
	Time series data has been used as well for profiling and
	failure detection~\cite{hauswirth2004vertical,
		hauswirth2005automating,sweeney2004using,adamoli2010trevis,
		abreu2007accuracy}.
	In particular, Hauswirth et al.~\cite{hauswirth2005automating}
	consider different performance metrics of
	a system such as instruction per cycle
	(IPC) and monitor these factors over time.
	Then, they apply dynamic time warping
	(DTW)~\cite{berndt1994using}
	to combine the traces of the same input.
	When they observe a pattern like sudden
	IPC changes, they align all traces
	using DTW and apply statistical
	correlation measurements to find
	predicates that are highly correlated with
	the changes in the target metric as the
	cause of performance anomaly.
	In our work, however, we do not collect
	metrics over time as a time series, although an extension is
	possible. As we collect only the total execution time, we do not need
	trace alignment, but we efficiently cluster traces
	based on relationships between running time
	and user inputs.
	
	\paragraph{Functional data clustering.}
	Functional data clustering is a technique
	for clustering given {\em functions}
	~\cite{jacques2014functional,
		jacques2014model,abraham2003unsupervised}.
	Even though there is a similarity in names,
	we are not given functions. Instead
	we are given individual data points,
	and we are discovering (linear) functions
	by clustering.
	Investigating the use of functional data
	clustering for debugging of software is
	an interesting area left for future work.

\bibliographystyle{plain}
\bibliography{papers}

\newpage
\section{Supplemental material A: Alignment Kernel}
\label{sec:appendix-A}
Figure~\ref{AlignmentKernel} shows the algorithm uses
to calculate the similarity matrix for spectral clustering based
on the notion of alignment kernel.
\begin{algorithm}[!th]\normalsize
{\scriptsize 
	\DontPrintSemicolon
	\KwIn{A set of points $\Ee = \set{\seq{\vx_1, \vy_1}, \ldots,
			\seq{\vx_n, \vy_n}}$, the number of desired clusters $k$,
		the number of nearest points $a$, minimum
		distance $\Delta$ between a point and function}
	\KwOut{A $n \times n$ similarity matrix represents a similarity graph.}
	initialize every element in similarity matrix $W_{n \times n}$ to $2.0$ and set $i$ to 0
	
	\While{$i < n $ }
	{Set $itr$ to 0
		
		\While{$itr < a $ }
		{
			let $rand$ be a random point from the neighbors of index point $i$.
			
			Set $j$ to $rand$
			
			\If{$i \neq j$ and $W[i][j] = 2.0$}
			{
				Fit a line passing through points $i$ and $j$.
				Let $M$ and $c$ be the slope vector and intercept of line.
				
				Calculate a set of points $R$.
				A point $r$ $\in$ $R$ ($0 \leq r < n$) if and only if $r \neq i$, $r \neq j$, and
				$| \vy_r - $ $(M^{T}$.$\vx_r$ $+$ $c) |$ $<$ $\Delta$
				
				\If{$|R|$ = 0 }
				{Set $W[i][j]$ and $W[j][i]$ to $Max\_Int$}
				\Else
				{
					Set $W[i][j]$ and $W[j][i]$ to  $\dfrac{1}{2^{|R|}}$
					
					\ForEach{$r$ in $R$} 
					{
						\If{$W[i][j]<W[i][r]$}
						{Set $W[i][r]$ and $W[r][i]$ to $W[i][j]$}
						\If{$W[i][j]<W[r][j]$}
						{Set $W[j][r]$ and $W[r][j]$ to $W[i][j]$}
					}
				}
			}
			\If{$i$ = $j$}
			{Set $W[i][j]$ to 0}
			Set $itr$ to $itr+1$
		}
		Set $i$ to $i+1$
	}
	For every entry $w$ in $W$, set $w$ to $e^{-w}$
	
	\Return $W$.
	\caption{\scriptsize \textsc{Alignment Kernel}}
	\label{AlignmentKernel}
}
\end{algorithm}

\section{Supplemental material B: Case Study}
\label{sec:appendix-B}
 In this section, we aim to provide detailed information
 about each case study.
 \subsection{Apache FOP}
   \begin{figure*}[!t]
 	\begin{minipage}[b]{0.4\textwidth}
 	\scalebox{0.6}{
 		\begin{tikzpicture}
 		\begin{axis}[xlabel=Image size (in bytes),ylabel= Time (s)]
 		\addplot[
 		visualization depends on={\thisrow{id}\as\myvalue},
 		scatter/classes={
 			0={mark=+,blue},
 			1={mark=+,red}
 		},
 		scatter, only marks,
 		scatter src=explicit symbolic]
 		table[x=size,y=mean,meta=label]
 		{data/fop.dat};
 		\end{axis}
 		\end{tikzpicture}
 	}
 	\end{minipage}\hfill
 	\begin{minipage}[b]{0.4\textwidth}
 		 	\scalebox{0.6}{
 		\begin{tikzpicture}[align=center,node distance=0.8cm,->,thick,
 		draw = black!60, fill=black!60]
 		\centering 
 		\pgfsetarrowsend{latex}
 		\pgfsetlinewidth{0.3ex}
 		\pgfpathmoveto{\pgfpointorigin} 
 		\pgfusepath{stroke}          
 		
 		\node[dtreenode,initial above,initial text={}] at (0,0) (l0)  {
 			ps.ImageEncodingHelper.\\encodeRenderImageWith-\\DirectColorModelAsRGB};
 		\node[dtreenode,below right=of l0] (l2)
 		{profile.\\ColorProfileUtil.\\
 			getICC\_Profile};
 		\node[dtreenode,below left=of l2] (l3) {inline.Line\\LayoutManager\\.handleOverflow};
 		
 		\node[dtreeleaf,bicolor={red and red and 0.99},below left=of l0] (l1) {};
 		\node[dtreeleaf,bicolor={blue and red and 0.99},below right=of l2]  (l4) {};
 		\node[dtreeleaf,bicolor={red and blue and 0.85},below left=of l3] (l5) {};
 		\node[dtreeleaf,bicolor={blue and red and 0.9},below right=of l3]  (l6) {};		
 		
 		\path[->]  (l0) edge  node [left,pos=0.4] {$< 1 ~~$} (l1);
 		\path  (l0) edge  node [right, pos=0.4] {$~~ \geq 1$} (l2);
 		\path  (l2) edge  node [left] {$\leq 1 ~~$} (l3);
 		\path  (l2) edge  node [right] {$~~ > 1$} (l4);
 		\path  (l3) edge  node [left] {$=0 ~~$} (l5);
 		\path  (l3) edge  node [right] {$~~ =1$} (l6);
 		\end{tikzpicture}
 	}
 	\end{minipage}
 	\caption{(Left). Clusters of FOP data set using Spectral Clustering
 		algorithm with alignment kernel. We set the number of
 		clusters to be 2.
 		(Right). Learning decision tree model to explain clusters in
 		Figure~\ref{fig:Clusters-FOP} based on FOP function calls.
 		Each outgoing arrow shows the frequency of function
 		calls}
 	\label{fig:Clusters-FOP}
 	\label{fig:Classification-FOP}
 \end{figure*}
 Apache FOP (Formatting Objects Processor) is
 a Java application that reads a formatting object (FO)
 tree such as XML file and renders the resulting pages
 to a specified output such as PDF and PS.
 For example, the output document can contain
 static areas that appear on every page, an external graphic,
 and a table.
 The external graphic can be PNG, JPEG, and SVG image
 files. Apache FOP includes many external libraries to fulfill
 different services and contains thousands of functions.

 We consider different input requests to apache FOP
 and record function calls, different features of input
 requests, and time of execution. 
 The inputs are in XML format and includes text, web
 addresses, embedded images, and so on.
 Our instrumentation records information of
 2,768 functions from different
 libraries involved in processing the request.

 Given the data set of different interactions
 with apache FOP Java application, we want
 to explain possible execution time
 differences based on the internal of the application.
 In other words, our goal is to explain timing
 differences among various requests
 based on program internals with considering
 the input features.

 Using \toolname,
 first, we notice that there is a relationship between
 execution time and the size of input images using visualization.
 As there is a non-linear relationship, we apply Spectral
 clustering with alignment kernel
 algorithm where we set $K$ to 2.
 Figure~\ref{fig:Clusters-FOP} (left) shows the result of
 clustering algorithm. The blue and red clusters
 represents
 0.00367 * $size^{1.177}$ + 69.918 and
 0.00291 * $size^{0.851}$ + 1738.2, respectively.
 The clustering step indicates that there are timing
 differences possibly coming
 from other properties beyond the size of input files.
 In particular, we are interested in knowing that what
 internal properties of FOP explain the blue and red
 clusters. We apply decision tree
 discriminant model that discriminates between
 clusters based on frequency of function calls. Figure~\ref{fig:Classification-FOP} (right)
 shows the result of decision tree model for FOP data set.
 For example, if (\texttt{ps.ImageEncodingHelper...ModelAsRGB}
 is called at least one time and \texttt{profile.ColorProfileUtil.getICC\_Profile}
 is called more than one time, then the data point
 is assigned to blue cluster. 
 The height of decision tree is 3 and the model
 uses two labels corresponding to two functions
 between time and size of embedded image as
 leaf models. The accuracy of decision tree model is
 96.4\% that shows the confidence to distinguish
 clusters based on the internal features of the
 application. It takes 14.4 seconds for \toolname
 to build this model.

 From the usefulness of these models,
 \toolname first identifies the differences in
 execution time models, and then, it uses
 decision tree algorithm to distinguish the
 clusters based on FOP function calls.
 Using the discriminant model,
 we observe that the function in the root node
 of discriminant model is not
 called for images with JPEG format. This means that
 JPEG images are assigned to red cluster. Another
 interesting observation is the role of compression
 methods of embedded images in distinguishing
 clusters.
 The function \texttt{profile.ColorProfileUtil.getICC$\_$Profile}
 is called one time for every inputs; however, it calls another
 time for PNG images with \textit{deflate} compression
 type:
  \begin{scriptsize}
 	\estiloJava
 	\begin{lstlisting}
 tryToExtractICCProfile() { ...
 	comp = imn.getAttribute("compression");
 	if ("deflate".equalsIgnoreCase(comp))
 	{
 		...
 		while (!decompresser.finished()){
 			...
 		}
 		...
 		iccProf = ColorProfileUtil.getICC_Profile(bos);
 		...
 	}
 }
 	\end{lstlisting}
 \end{scriptsize}
 As a result, if the embedded image is PNG with deflate
 compression, the input is assigned to blue cluster. Another
 possibility for a PNG file to assign to blue cluster is that
 the input file overflows the allowable size of output file.

\subsection{Charts4j: Plot Assistant Library}
charts4j is a Java chart library that enables developers
to pro-grammatically generate nearly all the charts available
in the Google Chart API.
Charts4j can be incorporated into any Internet enabled
Java or web application environment.

Give the data set of different interactions with the library
for plotting charts,
we want to explain possible execution time
differences based on the internal of Charts4j library
for different class of inputs.
Our goal is to explain timing models based on program
internals considering the number of data points in plots.

\begin{figure*}[!t]
	\centering
	\begin{minipage}[b]{0.4\textwidth}
		\centering
		 	\scalebox{0.6}{
		\begin{tikzpicture}
		\begin{axis}[xlabel=Number of data points,ylabel=
		Time (s)]
		\addplot[
		visualization depends on={\thisrow{id}\as\myvalue},
		scatter/classes={
			0={mark=+,blue},
			1={mark=+,red},
			2={mark=+,green!40!black}
		},
		scatter, only marks,
		scatter src=explicit symbolic]
		table[x=size,y=mean,meta=label]
		{data/charts4j.dat};
		\end{axis}
		\end{tikzpicture}
	}
	\end{minipage}\hfill
	\begin{minipage}[b]{0.4\textwidth}
		\centering
		\scalebox{0.6}{
		\begin{tikzpicture}[align=center,node distance=0.8cm,->,thick, draw = black!60, fill=black!60]
		\pgfsetarrowsend{latex}
		\pgfsetlinewidth{0.3ex}
		\pgfpathmoveto{\pgfpointorigin} 
		\pgfusepath{stroke}
		
		\node[dtreenode,initial above,initial text={}] at (0,0) (l0)  {charts4j.\\DataUtil.\\Scale};
		\node[dtreenode,below left=of l0] (l1)  {charts4j.\\PlotImpl.\\$<$init$>$};
		\node[dtreeleaf,bicolor={cadmiumgreen and blue and .9},
		below right=of l0] (l2) {};
		\node[dtreeleaf,color={blue},below left=of l1]  (l3) {};
		\node[dtreeleaf,tricolor={blue and red and cadmiumgreen and .3 and .9},
		below right=of l1]  (l4) {};
		
		\path  (l0) edge  node [left] {$= 0~~~$} (l1);
		\path  (l0) edge  node [right] {$~~~= 1$} (l2);
		\path  (l1) edge  node [left] {$= 0~~~$} (l3);
		\path  (l1) edge  node [right] {$~~~= 1$} (l4);
		\end{tikzpicture}
	}
	\end{minipage}
	\caption{(left) Clusters of different relationships between number of items
		for plotting in charts4j and time of execution using Spectral Clustering,
		(right) Learning decision tree after clustering data points for charts4j library.}
	\label{fig:clusters-charts4j}
	\label{fig:Classification-charts4j}
\end{figure*}
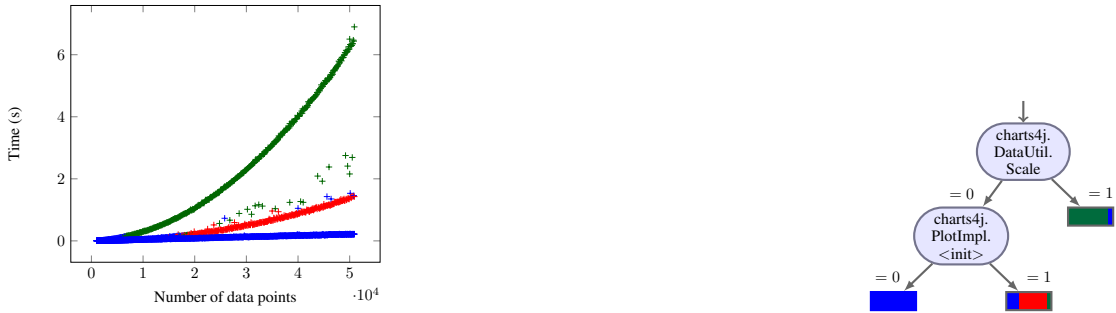

First, we notice that there is a relationship between
execution time and the number of items for plotting.
As there is a non-linear relationship, we apply Spectral
clustering with alignment kernel
algorithm where we set $K$ to 3.
Figure~\ref{fig:clusters-charts4j} (left) shows the result of
clustering algorithm.
The clustering step indicates that there are timing
differences possibly coming
from other properties beyond the number of items.
In particular, we are interested in knowing that what
internal properties of charts4j explain the blue,
red, and green clusters.
To explain the differences based
on program internals, we apply decision tree
discriminant model that distinguished between
clusters based on frequency of function calls.
Figure~\ref{fig:Classification-charts4j} (right)
shows the result of decision tree model for charts4j data set.
The height of decision tree is 2
and the model has three models in the leaf.
The accuracy of decision tree model is 87.1\% that
shows the confidence to distinguish clusters based
on the internal features of the application. It takes
11.7 seconds for \toolname to build this model.

CART regression tree builds a tree that has
the depth of 21, and it finds 1800 leaf models.
This suggests that CART is not a useful model
for our analysis. It takes 0.5 second for CART
to build this model, and coefficient of
determination ($R^2$) is 0.99.
M5Prime constructs a tree that has the depth
of 5, and it find 13 linear models in leafs.
This model is built in 17.3 seconds, and
$R^2$ is 0.99. GUIDE builds a similar tree
like \toolname. The tree has the depth of
2, and it finds three linear models in leaf.
It takes 1.5 seconds for GUIDE to build the
model, and $R^2$ is 0.97.

From the usefulness of these models,
\toolname gives us two important information:
first, we notice that green cluster is identified
by calling scale function that is not called for
other traces. Scale function takes an input,
find its minimum and maximum, and compute
the following for every element in the data set:
\texttt{scaledData[j]=((doubleArray[j] - min)/(max - min) * 100.0D);}.
Finally, it converts the new array to
double list. The class initialization for PlotImpl 
distinguishes blue and red clusters.
This happens for plots with one dimensional data
set where it generates new dimension for data
using the following loop:
\begin{scriptsize}
	\estiloJava
	\begin{lstlisting} 
	for (int i = 0; i < data.getSize(); i++)
	{
	xVals.add(Double.valueOf(x / 100.0D));
	x += inc;
	}
	this.xData = Data.newData(xVals);
	\end{lstlisting}
\end{scriptsize}
Finally, it calls newData() function to convert
the new data set to double array.
The obvious bug in this case is to convert
the list of double item to double type array
again. While, it can build double array structure
for new dimension. The following code shows
the part of application for converting to double type:
\begin{scriptsize}
	\estiloJava
	\begin{lstlisting} 
	double[] d = new double[data.size()];
	for (int i = 0; i < d.length; i++) {
	d[i] = ((Number)data.get(i)).doubleValue();
	}
	\end{lstlisting}
\end{scriptsize}

\subsection{SnapBuddy Social Network}
SnapBuddy is a Java web server that mocks a social network
in which each user has their own page with a public profile image.
It has 3,071 methods.

We instrument the application to obtain features
about the application's internals in every execution.
For collecting the data set,
we passively monitor different users' inputs with timing
information, users' input features (like size of file), and
internal function calls. The inputs are assumed to relate to
public profile of existing users in the system.
The instrumentation gives information about function calls'
frequency of 150 methods.
The goal is to detect possible timing differences in
downloading users' public profiles
and explain them based on the internals of SnapBuddy.
This analysis is also important from security point of view
where a malicious user can distinguish identity of users
who interact with the system through passively
observing execution time.

First, we apply M5Prime regression tree algorithm for the given
data set.
It detects 35 linear models in leafs and the
height of tree is 8. The coefficient of determination
is 0.99 that shows it predicts time with a high accuracy.
It takes 10 seconds for M5Prime to build the model.

Second, we apply GUIDE regression tree.
The regression tree uses 6 linear models and the height of tree is
3. The coefficient of determination for this model is 0.95.
It takes 1.7 seconds for GUIDE to build this model.

Third, we apply \toolname.
We plot execution time vs size of input files, and we observe
linear relationships between size of input and execution time.
We apply K-Linear clustering
algorithm where we set $K$ to 5.
Figure~\ref{fig:Clusters-SB} (left) shows
the result of clustering algorithm. For example, the blue
cluster is corresponding to the function
Time = 0.026 * size + 11.8
and the black cluster is the function
Time = 0.014 * size + 122.4.
The clustering step indicates that there
are timing differences possibly coming from some properties
of program internals that explain the differences.
We apply decision tree discriminant model
to distinguish between clusters based on program internal
features. Figure~\ref{fig:Classification-SB} (right)
shows the result of decision tree model for SnapBuddy data set.
For example, if (\texttt{$\lnot$KaleidoscopeFilter
	}$\land$\texttt{model.filter.Filter}$\land$
\texttt{TritoneFilter}$\land$
\texttt{EmbossFilter})$\lor$
(\texttt{KaleidoscopeFilter}$\land$
\texttt{$\lnot$TritoneFilter}$\land$
\texttt{$\lnot$EmbossFilter}) is true,
timing model is black cluster.
The height of decision tree is 4 and the model detects 5 lines.
The coefficient of determination for this model
is 0.99. The accuracy of decision tree model is 88.6\% that
shows the confidence to distinguish clusters based
on the internal features of the application. It takes 1.6
seconds for \toolname to build this model.

If we compare the results of GUIDE regression tree
and \toolname, we observe that
\toolname provides a complete
insight about the application internal behaviors that
accurately explain different timing models. In fact,
GUIDE misses to discover all of the relationships
between function calls and the functional clusters.

From the usefulness perspectives,
the discriminant model shows that the type
of image filters that users apply on their public
profile is the main reason for timing differences.
This can also lead to disclosure of a user's identity.
For example, if a malicious user observes the
interaction follows the red cluster in Figure~\ref{fig:Clusters-SB}
(left), she can guess that the user who interacts
with the system has no filter on his public profile image.
This observation can reduce the uncertainty of the
attacker about the identity of the user among all
users in the system.

\begin{figure*}[!t]
	\centering
	\begin{minipage}[b]{0.4\textwidth}
		\centering
		\scalebox{0.6}{
		\begin{tikzpicture}
		\begin{axis}[xlabel=Image size (in bytes),ylabel= Time (s)]
		\addplot[
		visualization depends on={\thisrow{id}\as\myvalue},
		scatter/classes={
			0={mark=+,cyan},
			1={mark=+,black},
			2={mark=+,red},
			3={mark=+,blue},
			4={mark=+,green!40!black}
		},
		scatter, only marks,
		scatter src=explicit symbolic]
		table[x=size,y=mean,meta=label]
		{data/SB.dat};
		\end{axis}
		\end{tikzpicture}
	}
	\end{minipage}\hfill
	\begin{minipage}[b]{0.4\textwidth}
		\scalebox{0.6}{
			              	\begin{tikzpicture}[align=center,node distance=1cm,->,thick,
			draw = black!60, fill=black!60]
			\pgfsetarrowsend{latex}
			\pgfsetlinewidth{0.3ex}
			\pgfpathmoveto{\pgfpointorigin} 
			\pgfusepath{stroke}
			
			\node[dtreenode,initial above,initial text={}] at (0,0) (l0)  {Image.\\~~~KaleidoscopeFilter.filter~~~};
			\node[dtreenode,below left=of l0] (l1) {model.\\filter.\\filter};
			\node[dtreenode,below right=of l0] (l2) {Image.\\~~~TritoneFilter.filter~~~};
			\node[dtreeleaf,bicolor={red and cadmiumgreen and 0.9},below left=of l1] (l3) {};
			\node[dtreenode,below=of l1] (l4)
			{image.\\Emboss\\Filter.\\filterPixel};
			\node[dtreenode,below left=of l2] (l5)
			{image.\\Emboss\\Filter.filterPixel};
			\node[dtreenode,below=3.5cm of l2] (l6)
			{image.\\Emboss\\Filter.filterPixel};
			\node[dtreeleaf,bicolor={cadmiumgreen and cyan and 0.9},below left=of l4] (l7) {};
			\node[dtreenode,below=of l4] (l8)
			{image.\\TritoneFilter.\\filter};
			\node[dtreeleaf,bicolor={black and blue and 0.8},below=of l5] (l9) {};
			\node[dtreeleaf,bicolor={cyan and blue and 0.8},below right=of l5] (l10) {};
			\node[dtreeleaf,bicolor={cyan and blue and 0.9},below=of l6] (l11) {};
			\node[dtreeleaf,bicolor={blue and cyan and 0.95},below right=of l6] (l12) {};
			\node[dtreeleaf,bicolor={cadmiumgreen and cyan and 0.9},below left=of l8] (l13) {};
			\node[dtreeleaf,bicolor={black and blue and 0.95},below right=of l8] (l14) {};	
			
			\path  (l0) edge  node [left] {$False ~~$} (l1);
			\path  (l0) edge  node [right] {$~~ True$} (l2);
			\path  (l1) edge  node [left] {$False ~~$} (l3);
			\path  (l1) edge  node [right] {$~~ True$} (l4);
			\path  (l2) edge  node [left] {$False ~~$} (l5);
			\path  (l2) edge  node [right] {$~~ True$} (l6);
			\path  (l4) edge  node [left] {$False ~~$} (l7);
			\path  (l4) edge  node [right] {$~~ True$} (l8);
			\path  (l5) edge  node [left] {$False ~~$} (l9);
			\path  (l5) edge  node [right] {$~~ True$} (l10);
			\path  (l6) edge  node [left] {$False ~~$} (l11);
			\path  (l6) edge  node [right] {$~~ True$} (l12);
			\path  (l8) edge  node [left] {$False ~~$} (l13);
			\path  (l8) edge  node [right] {$~~ True$} (l14);
			\end{tikzpicture}
		}
	\end{minipage}
	\caption{(Left) Clusters of SnapBuddy data set where each cluster is a distinct linear function that explains time based on size of profile image. (Right) Learning decision tree model to explain clusters in Figure~\ref{fig:Clusters-SB} based on SnapBuddy function calls.}
	\label{fig:Clusters-SB}
	\label{fig:Classification-SB}
\end{figure*}

\subsection{JFreeChart}
JFreeChart is a free Java chart library that
makes it easy for developers to display
professional quality charts in their applications.

Give the data set of open-high-low-close (OHLC),
we collect time of execution
for plotting an input, number of items for plotting,
and frequency of function calls in JFreechart
library. We intend to detect possible timing
differences and pinpoint parts of library responsible
for the differences.

We apply \toolname to first plot execution time
vs number of items in OHLC dataset.
We apply K-linear clustering algorithm where
we set $K$ to 2, and it finds two linear
clusters shown in Figure~\ref{fig:clusters-jfreechart} (left).
To explain the differences based
on program internals, we apply decision tree discriminant model
that distinguishes between clusters based on program internal features. Figure~\ref{fig:Classification-jfreechart} (right)
shows the result of decision tree model that
explains differences between red and blue clusters.

We observe that whether the plot applies
\texttt{CandlestickRenderer} and calls
\texttt{drawItem} distinguishes
between red and blue clusters. 
%Looking into
%this method, we can see a condition that leads
%to execution of $O(n)$ loop:\\
% \begin{scriptsize}
%	\estiloJava
%	\begin{lstlisting}
%	if(this.candleWidth = 0.0D)
%		switch (this.autoWidthMethod)
%		{
%			case 0:
%			... // no loop statements
%			break;
%			case 1: 
%				itemCount = highLowData.getItemCount(series);
%				for (int i = 0; i < itemCount; i++)
%				{
%			 	...
%			 	}
%			break;
%			case 2: 
%				... // no loop statements
%		}
%	\end{lstlisting}
%\end{scriptsize} 
In particular, our model shows that the number of
calls to \texttt{drawItem} is equal to number of data
points in OHLC dataset, and each call executes
the loop statement above in the order of number
of data item if the conditions are satisfied. This
is a potential performance problem that is also
presented as a performance bug in previous works.

\begin{figure*}[!t]
	\centering
	\begin{minipage}[b]{0.4\textwidth}
		\centering
		 	\scalebox{0.6}{
		\begin{tikzpicture}
		\begin{axis}[xlabel=Number of data points,ylabel=
		Time (s)]
		\addplot[
		visualization depends on={\thisrow{id}\as\myvalue},
		scatter/classes={
			0={mark=+,blue},
			1={mark=+,red}
		},
		scatter, only marks,
		scatter src=explicit symbolic]
		table[x=size,y=mean,meta=label]
		{data/jfreechart.dat};
		\end{axis}
		\end{tikzpicture}
	}
	\end{minipage}\hfill
	\begin{minipage}[b]{0.4\textwidth}
		\centering
		\scalebox{0.6}{
		\begin{tikzpicture}[align=center,node distance=2.2cm,->,thick, draw = black!60, fill=black!60]
		\pgfsetarrowsend{latex}
		\pgfsetlinewidth{0.3ex}
		\pgfpathmoveto{\pgfpointorigin} 
		\pgfusepath{stroke}
		
		\node[dtreenode,initial above,initial text={}] at (0,0) (l0)  {xy.Candlestick\\Renderer.drawItem};
		\node[dtreeleaf,bicolor={blue and red and .97},
		below left=of l0] (l1) {};
		\node[dtreeleaf,bicolor={red and blue and .95},
		below right=of l0] (l2) {};
		
		\path  (l0) edge  node [left] {$= 0 ~~$} (l1);
		\path  (l0) edge  node [right] {$~~> 0$} (l2);
		\end{tikzpicture}
	}
	\end{minipage}
	\caption{(Left) Clusters of different relationships between number of items for plotting in jfreechart
	and time of execution using K-linear Clustering,
	(Right) Learning decision tree after clustering data points for jfreechart library.}
	\label{fig:clusters-jfreechart}
	\label{fig:Classification-jfreechart}
\end{figure*}

\subsection{Apache POI}
The Apache POI Project's mission is to read and
write MS Excel, MS Word, and
PowerPoint files using Java.
A major use of the
Apache POI is for Text Extraction applications
such as web spiders, index builders, and content
management systems. We would use HSLF API
for MS PowerPoint.

Given data set of different slides randomly
included text, table, images, shapes, and smartArts,
we collect time taken to convert
slides to images, the size of slides,
and frequency of function calls in the Apache POI
library. We intend to detect possible timing
differences and pinpoint parts of library responsible
for the differences.

We apply Spectral clustering algorithm where
we set $K$ to 3, and it finds three
clusters shown in Figure~\ref{fig:clusters-poi}
(left). Note that images and special items can't
exceed the size of slides. This can be seen for
green and red clusters that can't exceed 170 KB.
To explain the differences based
on program internals, we apply decision tree
discriminant model that distinguishes between
clusters based on program internal features.
Figure~\ref{fig:Classification-poi} (right)
shows the result of decision tree model that
explains differences between blue, green, and red
clusters.

The decision tree model indicates that all traces
labeled with blue cluster do not have any image
data source. These include pictures and smartArts.
Then, the number of Escher records in an item distinguishes
green and red clusters. Based on the experiments,
slides with PNG images tend to follow red cluster,
while slides with smartArts follow the green cluster.
All other materials like table, shapes, and texts follow the
blue cluster.

\begin{figure*}[!t]
	\centering
	\begin{minipage}[b]{0.4\textwidth}
		\centering
		\scalebox{0.6}{
		\begin{tikzpicture}
		\begin{axis}[xlabel=Size of slides in Byte,ylabel=
		Time (s)]
		\addplot[
		visualization depends on={\thisrow{id}\as\myvalue},
		scatter/classes={
			0={mark=+,blue},
			1={mark=+,red},
			2={mark=+,green!40!black}
		},
		scatter, only marks,
		scatter src=explicit symbolic]
		table[x=size,y=mean,meta=label]
		{data/POI.dat};
		\end{axis}
		\end{tikzpicture}
	}
	\end{minipage}\hfill
	\begin{minipage}[b]{0.4\textwidth}
		\centering
		\scalebox{0.6}{
		\begin{tikzpicture}[align=center,node distance=0.8cm,->,thick, draw = black!60, fill=black!60]
		\pgfsetarrowsend{latex}
		\pgfsetlinewidth{0.3ex}
		\pgfpathmoveto{\pgfpointorigin} 
		\pgfusepath{stroke}
		
		\node[dtreenode,initial above,initial text={}] at (0,0) (l0)  {hslf.blip.\\PNG.getType};
		\node[dtreeleaf,color={blue}, below left=of l0] (l1) {};
		\node[dtreenode,below right=of l0] (l2) {ddf.EscherSimple\\Property.getPropertyValue};
		\node[dtreeleaf,bicolor={green!40!black and blue and .7},below left=of l2] (l3) {};
		\node[dtreeleaf,color={red},below right=of l2] (l4) {};
		
		\path  (l0) edge  node [left] {$= 0 ~~$} (l1);
		\path  (l0) edge  node [right] {$~~> 0$} (l2);
		\path  (l2) edge  node [left] {$<= 20 ~~$} (l3);
		\path  (l2) edge  node [right] {$~~> 20$} (l4);
		\end{tikzpicture}
	}
	\end{minipage}
	\caption{(Left) Clusters of different relationships
		between the size of slide
		and time of execution using Spectral Clustering,
		(Right) Learning decision tree after clustering
		data points.}
	\label{fig:clusters-poi}
	\label{fig:Classification-poi}
\end{figure*}

\subsection{Ode4j}
Ode4j is an open source, high performance library
for simulating rigid body dynamics.
It has advanced joint types and integrated
collision detection with friction.
Ode4j is useful for simulating vehicles, objects
in virtual reality environments and virtual creatures.

Given different number of geom objects,
we use different APIs to evaluate their timing
performance and explain possible timing model
differences based on the application's internals.

We apply Spectral clustering algorithm where
we set $K$ to 2, and it finds two
clusters shown in Figure~\ref{fig:clusters-ode4j} (left).
To explain the differences based
on program internals, we apply decision tree
discriminant model.
Figure~\ref{fig:Classification-ode4j} (right)
shows the result of decision tree model that
explains differences between red and blue
clusters.

The discriminant model detects \texttt{collide}
function in \texttt{DxSAPSpace} class as a feature
to distinguish blue and red clusters.
This function is called one or more times for red cluster,
while it does not call for blue cluster. Looking into source
code, we can see the quadratic behavior in
\texttt{collide} function. As there is a linear cost function
named \texttt{collide2} in \texttt{DxSAPSpace} class,
it depends on developers to decide if this behavior can
be optimized or not.  

\begin{figure*}[!t]
	\centering
	\begin{minipage}[b]{0.4\textwidth}
		\centering
		 	\scalebox{0.6}{
		\begin{tikzpicture}
		\begin{axis}[xlabel=Number of geom objects,ylabel=
		Time (s)]
		\addplot[
		visualization depends on={\thisrow{id}\as\myvalue},
		scatter/classes={
			0={mark=+,blue},
			1={mark=+,red}
		},
		scatter, only marks,
		scatter src=explicit symbolic]
		table[x=size,y=mean,meta=label]
		{data/ode4j.dat};
		\end{axis}
		\end{tikzpicture}
	}
	\end{minipage}\hfill
	\begin{minipage}[b]{0.4\textwidth}
		\centering
		 	\scalebox{0.6}{
		\begin{tikzpicture}[align=center,node distance=2.2cm,->,thick, draw = black!60, fill=black!60]
		\pgfsetarrowsend{latex}
		\pgfsetlinewidth{0.3ex}
		\pgfpathmoveto{\pgfpointorigin} 
		\pgfusepath{stroke}
		
		\node[dtreenode,initial above,initial text={}] at (0,0) (l0)  {internal.DxSAP.\\Space.collide};
		\node[dtreeleaf,bicolor={blue and red and 0.97}, below left=of l0] (l1) {};
		\node[dtreeleaf,bicolor={red and blue and 0.75},below right=of l0] (l2) {};
		
		\path  (l0) edge  node [left] {$= 0 ~~$} (l1);
		\path  (l0) edge  node [right] {$~~> 0$} (l2);
		\end{tikzpicture}
	}
	\end{minipage}
	\caption{(Left) Clusters of different relationships
		between the number of geom objects in ode4j
		and time of execution using Spectral clustering,
		(Right) Learning decision tree after clustering
		data points.}
	\label{fig:clusters-ode4j}
	\label{fig:Classification-ode4j}
\end{figure*}

\subsection{Collab}
Collab is a scheduling application which allows
users to create new events and modify existing
events. Each user's event is identified by a unique
event ID which represents the number of time units
since the application’s epoch.

Given a data set of different operations to add and
modify events using \texttt{ADD},\texttt{UNDO},
and \texttt{DONE}, we want to analyze possible
timing differences and explain the differences based
on program internal features.

We apply K-linear clustering algorithm where
we set $K$ to 2, and it finds two
clusters shown in Figure~\ref{fig:clusters-collab} left).
To explain the differences based
on program internals, we apply decision tree
discriminant model.
Figure~\ref{fig:Classification-collab} (right)
shows the result of decision tree model that
explains differences between blue and red
clusters.

The discriminant model tells us that the number
of calls to \texttt{split} function distinguishes red and
blue clusters. It says that traces in red cluster
calls \texttt{split} function more times than traces in
blue cluster. In fact, the average number of calls to
\texttt{split} function for blue cluster is 1,678 and for
red cluster is 88,507. 

Looking into \texttt{SchedulingSandBox},
we can see the following code snippet that calls 
\texttt{split} function:
 \begin{scriptsize}[!t]
	\estiloJava
	\begin{lstlisting} 
	if (eventidlist[ind] == -2)
	{
		this.log.publish("node full");
		...
		split(node);
		addhelper(lval);
	}
	\end{lstlisting}
\end{scriptsize}

The sandbox class uses B-tree structure to
store nodes temporary.
The discriminant model suggests the number
of times that B-tree needs to split nodes is an important
feature to distinguish red and blue clusters.
In the implementation, after adding 8 nodes,
B-tree requires to split over the parent node.
If the user commits exactly after 8 additions, this will lead
to the highest number of split required by B-tree. 
One possible solution is to increase the capacity of
each individual node in B-tree and require users' to
commit after a certain number of \texttt{ADD} operations.

\begin{figure*}[!t]
	\centering
	\begin{minipage}[b]{0.4\textwidth}
		\centering
		 	\scalebox{0.6}{
		\begin{tikzpicture}
		\begin{axis}[xlabel=Number of ADD operations,ylabel=
		Time (s)]
		\addplot[
		visualization depends on={\thisrow{id}\as\myvalue},
		scatter/classes={
			0={mark=+,blue},
			1={mark=+,red}
		},
		scatter, only marks,
		scatter src=explicit symbolic]
		table[x=size,y=mean,meta=label]
		{data/collab.dat};
		\end{axis}
		\end{tikzpicture}
	}
	\end{minipage}\hfill
	\begin{minipage}[b]{0.4\textwidth}
		\centering
		 	\scalebox{0.6}{
		\begin{tikzpicture}[align=center,node distance=2.2cm,->,thick, draw = black!60, fill=black!60]
		\pgfsetarrowsend{latex}
		\pgfsetlinewidth{0.3ex}
		\pgfpathmoveto{\pgfpointorigin} 
		\pgfusepath{stroke}
		
		\node[dtreenode,initial above,initial text={}] at (0,0) (l0)  {collab.Scheduling\\Sandbox.split};
		\node[dtreeleaf,bicolor={blue and red and 0.97}, below left=of l0] (l1) {};
		\node[dtreeleaf,bicolor={red and blue and 0.98}, below right=of l0] (l2) {};
		\path  (l0) edge  node [left] {$<= 2805~~$} (l1);
		\path  (l0) edge  node [right] {$~~> 2805$} (l2);
		\end{tikzpicture}
	}
	\end{minipage}
	\caption{(Left) Clusters of different relationships
		between the number of ADD operations
		and time of execution using K-Linear,
		(Right) Learning decision tree after clustering
		data traces in Collab.}
	\label{fig:clusters-collab}
	\label{fig:Classification-collab}
\end{figure*}

\subsection{Tweeter}
Tweeter (formerly spell corrector) application
is a mock of twitter. Users can create a profile,
post tweets using hash-tags,
and see tweets posted by other users.

Given tweets of maximum 20 words,
we want to see how long it takes for tweeter
to post the tweet, and we intend to discover
and explain potential timing differences based on
application's internal.

We apply K-linear clustering algorithm where
we set $K$ to 2, and it finds two
clusters shown in Figure~\ref{fig:clusters-tweeter}
(left).
To explain the differences based
on program internals, we apply decision tree
discriminant model.
Figure~\ref{fig:Classification-tweeter} (right)
shows the result of decision tree model that
explains differences between red and blue
clusters.

The discriminant model introduces \texttt{getAlternatives}
function as a feature to distinguish blue and red clusters.
This function is called more times when the input tweet
includes mistakes, and it will never be called if the tweet
is correct:
 \begin{scriptsize}
	\estiloJava
	\begin{lstlisting} 
	for(String word: tweet) {
		...
		final Map<.,.> correct = spelling.correct(word);
		if (correct != null) {
			...
			getAlternatives().add(corr);
		}
		...
	}
	\end{lstlisting}
\end{scriptsize} 
\texttt{getAlternatives} collects all possible
prefixes to match the words in the tweet with correct
words in vocabulary.
As a result, longer tweets with multiple mistakes would
need more time to process.

\begin{figure*}[!t]
	\centering
	\begin{minipage}[b]{0.4\textwidth}
		\centering
		\scalebox{0.6}{
		\begin{tikzpicture}
		\begin{axis}[xlabel=Number of words in a tweet,ylabel=
		Time (s)]
		\addplot[
		visualization depends on={\thisrow{id}\as\myvalue},
		scatter/classes={
			0={mark=+,blue},
			1={mark=+,red}
		},
		scatter, only marks,
		scatter src=explicit symbolic]
		table[x=size,y=mean,meta=label]
		{data/tweeter.dat};
		\end{axis}
		\end{tikzpicture}
	}
	\end{minipage}\hfill
	\begin{minipage}[b]{0.4\textwidth}
		\centering
		 	\scalebox{0.6}{
		\begin{tikzpicture}[align=center,node distance=2.2cm,->,thick, draw = black!60, fill=black!60]
		\pgfsetarrowsend{latex}
		\pgfsetlinewidth{0.3ex}
		\pgfpathmoveto{\pgfpointorigin} 
		\pgfusepath{stroke}
		
		\node[dtreenode,initial above,initial text={}] at (0,0) (l0)  {hibernate.Alternative\\Suggestions.\\getAlternatives};
		\node[dtreeleaf,color={blue}, below left=of l0] (l1) {};
		\node[dtreeleaf,color={red},below right=of l0] (l2) {};
		
		\path  (l0) edge  node [left] {$= 0 ~~$} (l1);
		\path  (l0) edge  node [right] {$~~> 0$} (l2);
		\end{tikzpicture}
	}
	\end{minipage}
	\caption{(Left) Clusters of different relationships
		between the number of words in tweets
		and time of execution using K-Linear clustering,
		(Right) Learning decision tree after clustering
		data points.}
	\label{fig:clusters-tweeter}
	\label{fig:Classification-tweeter}
\end{figure*}

\end{document}